\documentclass[10pt,twocolumn,letterpaper]{article}

\usepackage{wacv}              

\usepackage{graphicx}
\usepackage{amsmath}
\usepackage{amssymb}
\usepackage{booktabs}
\usepackage{multirow}
\usepackage{makecell}
\usepackage[table,xcdraw]{xcolor}
\usepackage{array}
\usepackage{ragged2e}
\usepackage{colortbl}
\usepackage{tabularray}
\usepackage{color}

\usepackage{pifont}
\usepackage[normalem]{ulem}
\usepackage{xcolor}
\newcommand{\cmark}{{\color{green}\ding{51}}}
\newcommand{\xmark}{{\color{red}\ding{55}}} 

\usepackage[capitalize]{cleveref}
\crefname{section}{Sec.}{Secs.}
\Crefname{section}{Section}{Sections}
\Crefname{table}{Table}{Tables}
\crefname{table}{Tab.}{Tabs.}

\def\wacvPaperID{803} 
\def\confName{WACV}
\def\confYear{2026}

\begin{document}

\title{Diagnose Like A REAL Pathologist: An Uncertainty-Focused Approach for Trustworthy Multi-Resolution Multiple Instance Learning}

\author{Sungrae Hong, Sol Lee, Jisu Shin, Jiwon Jeong, Mun Yong Yi\thanks{Corresponding Author}\\
Korea Advanced Institute of Science and Technology, Daejeon, South Korea\\
{\tt\small \{sr5043, leesol4553, jisu3389, zzioni, munyi\}@kaist.ac.kr}
}



\maketitle


\begin{abstract}
    With the increasing demand for histopathological specimen examination and diagnostic reporting, Multiple Instance Learning (MIL) has received heightened research focus as a viable solution for AI-centric diagnostic aid. Recently, to improve its performance and make it work more like a pathologist, several MIL approaches based on the use of multiple-resolution images have been proposed, delivering often higher performance than those that use single-resolution images. {Despite impressive recent developments of multiple-resolution MIL, previous approaches only focus on improving performance, thereby lacking research on well-calibrated MIL that clinical experts can rely on for trustworthy diagnostic results.} In this study, we propose Uncertainty-Focused Calibrated MIL (UFC-MIL), which more closely mimics the pathologists’ examination behaviors while providing calibrated diagnostic predictions, using multiple images with different resolutions. UFC-MIL includes a novel patch-wise loss that learns the latent patterns of instances and expresses their uncertainty for classification. Also, the attention-based architecture with a neighbor patch aggregation module collects features for the classifier. In addition, aggregated predictions are calibrated through patch-level uncertainty without requiring multiple iterative inferences, which is a key practical advantage. Against challenging public datasets, UFC-MIL shows superior performance in model calibration while achieving classification accuracy comparable to that of state-of-the-art methods.
\end{abstract}

\begin{figure}[t]
  \centering
   \includegraphics[width=0.8\linewidth]{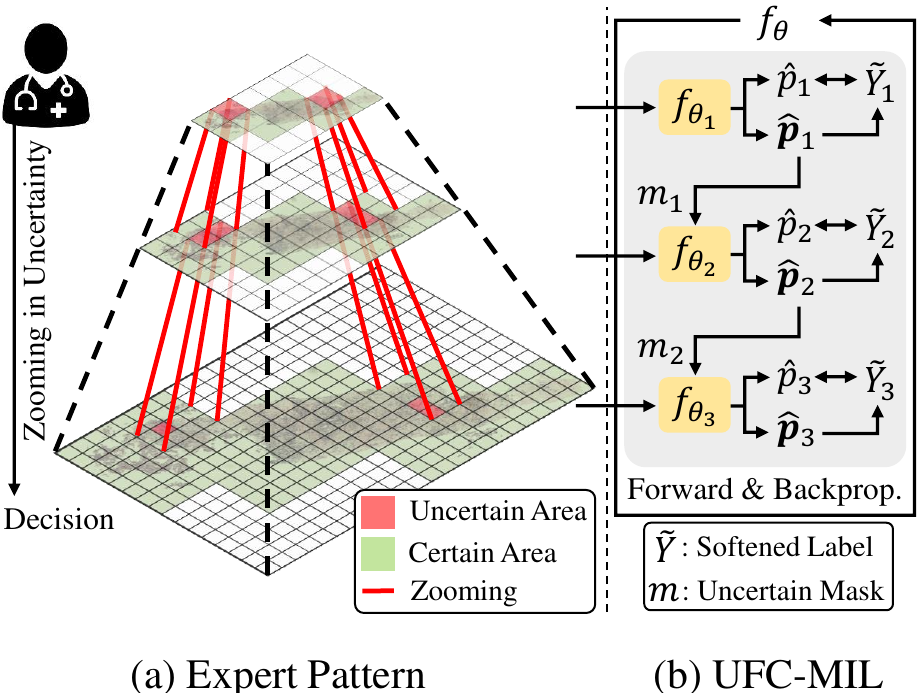}
   \caption{An illustration of the pathologists' observation pattern and the key mechanisms of UFC-MIL that reflect the pattern. (a) Pathologists begin observation at the coarsest resolution, identifying uncertain areas for further scrutiny. They zoom into these area to acquire additional information for diagnosis. (b) UFC-MIL, equipped with multi-resolution patches, focuses on sub-patches of those identified as uncertain at higher resolutions. Patch-level uncertainty at each resolution is then applied to calibration.}
   \label{fig:1}
\end{figure}
\section{Introduction}
The post-COVID-19 era has seen an explosion in demand for pathological diagnoses, placing an untenable burden on the constrained number of pathologists~\cite{article,bray2021ever}. AI-based models present a feasible solution to help pathologists and relieve their burden; however, annotating megapixels of whole slide images (WSI) significantly increases pathologist workload~\cite{zhang2025patches}. Consequently, the deep learning (DL) community is vigorously exploring Multiple Instance Learning (MIL), a weakly supervised approach that requires only WSI-level labels to classify pathological images~\cite{gadermayr2024multiple}.


Previous research on MIL has been mainly concerned with developing histopathological diagnosis models on the basis of a single resolution WSI~\cite{ilse2018attention,shao2021transmil,zhang2022dtfd,lu2021data}. Recently, recognizing that pathologists consult multiple resolutions for diagnosis, multi-resolution MIL (MRMIL) has gained increased attention producing improved performance, as the approach seeks to exploit richer and more fine-grained details~\cite{gadermayr2024multiple}. MRMIL models observe WSIs in all available resolutions utilizing graphs~\cite{hou2022h,bontempo2023mil}, image pyramids~\cite{li2021dual}, and entire patches\footnote{For clarity, we use "instance" and "patch" interchangeably.}~\cite{chen2022scaling,huang2023cross,xiong2023diagnose}.

The primary objective of pathology MIL is to help clinical specialists by screening diagnoses~\cite{quellec2012multiple}. Thus, it is crucial for MIL models to produce interpretable and acceptable results to end-user clinicians~\cite{eloy2023artificial}. As deep learning networks tend to become overconfident with increasing depth, training without model calibration leads to biased confidence results that diverge from human judgment~\cite{guo2017calibration}, which has critical implications in medical diagnosis. Overestimating Type II (false negative) errors can deprive patients of timely treatment opportunities~\cite{raab2005clinical}. Repeated instances of incorrect predictions due to the overestimation of MIL models, in the long run, undermine the reliance of clinicians on their output~\cite{dolezal2022uncertainty}. For these reasons, it is imperative that MIL research shifts its current dominant focus from improving performance to improving calibration to make it easily applicable to clinical settings. 

Pathologists' behavioral patterns also provide important intuition for the development of MIL models that are practically relevant. They begin with {the coarsest resolution}, then zoom in on specific areas requiring more focused observation to examine finer resolutions~\cite{brunye2017accuracy} as shown in Fig.~\ref{fig:1}. Their determination of regions of interest is not driven by mere randomness, but by the need for focused observation of uncertain areas {critical to a diagnosis}~\cite{ghezloo2022analysis}. In other words, the process of pathologists' zooming resolves diagnostic uncertainty arising at a coarser level by increasing the amount of information. The issue of uncertainty, in turn, reverts to the MIL calibration.

Although MRMIL shows impressive performance, there is a gap in achieving a well-calibrated model for clinical applications. Toward overcoming this limitation, we propose Uncertainty-Focused Calibrated MIL (UFC-MIL), which emulates multi-resolution expert observation patterns while simultaneously addressing the neglected calibration issue. UFC-MIL includes a novel patch-wise loss to measure per-instance uncertainty by generating individual predictions. This term enables the model to learn individual instance judgments against weak labels without violating the fundamental MIL assumption. The expert's uncertainty-driven zooming pattern is simulated by a differentiable mask generation and the cross-attention module, identifying instances with high entropy and allowing the model to deliver features that facilitate a more focused examination of their subinstances. Taking into account the spatial invariance characteristic of pathology images, the Topological Neighbor Attention Module (TNAM) in UFC-MIL aggregates information from neighboring patches for individual patches. Furthermore, its model calibration solution introduces Sample and Resolution-wise Label Smoothing (SRLS). This accounts for the varied information content and heterogeneous uncertainty between resolutions and samples.

We summarize our contribution as follows.
\begin{itemize}
    \item{We propose UFC-MIL, which mimics the top-down zooming behaviors of medical experts in uncertain areas, and simultaneously introduces a calibration method that leverages its output structure. To our knowledge, this is the first attempt to address the calibration issue of MRMIL.}
    \item{Components of UFC-MIL enable end-to-end training of multi-resolution WSIs. Specifically, the proposed patch-wise loss allows for patch-level predictions, which can then be utilized for calibration, without violating the MIL assumption.}
    \item{Experiments conducted extensively on public datasets demonstrate that UFC-MIL, combined with the proposed SRLS, an inference-free calibration training approach that leverages multiple outputs, shows superior performance in model calibration and exhibits classification performance comparable to that of state-of-the-art MRMIL architectures.}
\end{itemize}

\section{Related Work}
\subsection{Multi-Resolution Multiple Instance Learning}
Multiple Instance Learning (MIL) assumes that a WSI $X_i$ is a bag $\{x_1, \cdots, x_{n}\}$, where each $x_n$ is defined as a valid patch from $X_i$. Only the WSI level label $Y_i$ is given: 
\begin{equation}
    Y_i = 
    \begin{cases}
    0 \text{ , iff } \sum_{n} y_{n} = 0 \\
    1 \text{ , otherwise}
    \end{cases}
\end{equation}
where $X_i$ is considered negative if all of its instances are negative but is positive if at least one patch is positive. The pre-trained feature extractor maps all instances to a low-dimensional space: $x_n\rightarrow{z_n}\in\mathbb{R}^d$, where $d$ is the dimension of the feature. MIL aggregator merges individual instance features to predict the label $\hat{Y}_i$.

The early MIL implementations incorporated hand-made maximum, minimum, and mean aggregators~\cite{ramon2000multi}. The rise of attention-based models allowed MIL aggregators to yield more explainable results~\cite{ilse2018attention,shao2021transmil}. 
DTFD-MIL~\cite{zhang2022dtfd} adeptly captured subtle data information via a double-tier mechanism that partitions instances into several pseudo bags. 
{Meanwhile, highlighting the ambiguity of attention-based evidence, xMIL~\cite{hense2024xmil} proposed Layer-wise Relevance Propagation (LRP) to compute instance influence. ProtoMIL~\cite{rymarczyk2022protomil} introduced a prototype layer to cluster positive and negative classes based on instance bag similarity.} However, these methods only use the given information in a limited single-resolution way.

\begin{figure*}[t]
  \centering
  \includegraphics[width=0.92\textwidth]{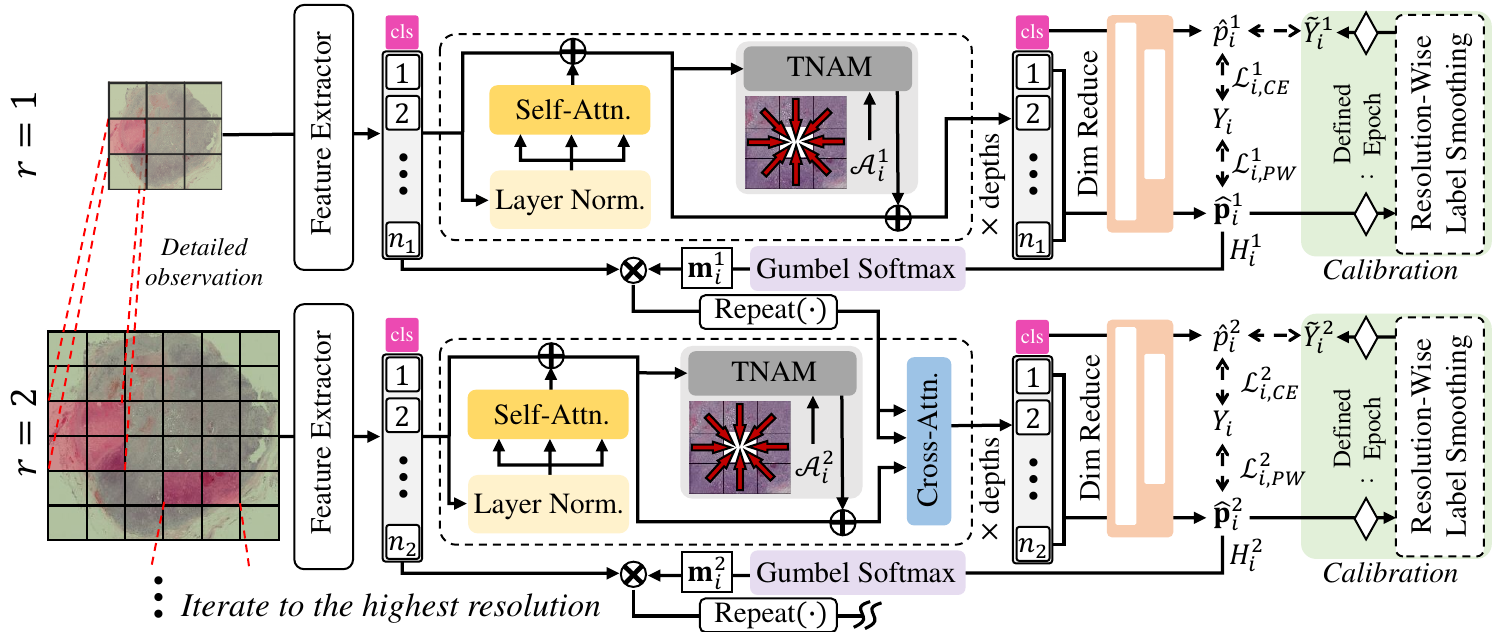} 
  \caption{Overview of UFC-MIL, which employs a top-down analysis from the coarsest $(r=1)$ to the finest $(r=R>1)$ resolution.}
  \label{fig:ufc-mil}
\end{figure*}

The DL community has recently been leveraging the rich representations from multi-resolution WSI. DS-MIL~\cite{li2021dual} uses multiresolution instances by concatenating features into image pyramids. Graph-based methods compress patch structure information through message passing and aggregation~\cite{bontempo2023mil,hou2022h}. HIPT~\cite{chen2022scaling}, with its Vision Transformer~\cite{dosovitskiy2020image} architecture, observes all patches from fine to coarse levels. Xiong et al.~\cite{xiong2023diagnose} highlighted that the employment of thousands to tens of thousands of instances per sample differs from typical pathologist behavior and risks introducing redundant information. Despite remarkable WSI classification performance, these methods do not align well with humans' WSI investigation behaviors, particularly the top-down zooming in behaviors for closer examination of uncertain areas.

\subsection{Calibration for MIL}

Model calibration comprises post-hoc, regularization, and uncertainty estimation methods~\cite{wang2023calibration}. Post-hoc methods~\cite{zadrozny2001obtaining,guo2017calibration} calibrate the model using hyperparameters selected from a validation set. However, their performance is sensitive to the validation set and the choice of hyperparameters. Regularization-based methods~\cite{pereyra2017regularizing,muller2019does} improve model calibration through learnable parameters, although hyperparameter selection remains crucial for effective training. Uncertainty estimation approaches~\cite{blundell2015weight,gal2016dropout,lakshminarayanan2017simple} infer network statistics through iterative learning and repeated inference, which can lead to time restrictions in practical applications, such as MIL.

Few works have proposed calibration methods specifically for reliable MIL. Park et al.~\cite{park2025uncertainty} introduced Uncertainty-based Data-wise Label Smoothing (UDLS), a sample-wise calibration training, positing that individual WSIs have differing uncertainties. In their study, sample-specific uncertainty is obtained by inferring the model after randomly dropping patch features multiple times. The accumulated uncertainties across all samples subsequently guide the retraining of the entire model through label smoothing. However, this approach requires dozens of iterative inference steps, unlike our method. Furthermore, model calibration in multi-resolution MIL is an unexplored avenue.
\section{Method}
We propose UFC-MIL, which mimics expert behavior patterns, and simultaneously introduce its applicable calibration training (Fig.~\ref{fig:ufc-mil}). 


\subsection{End-to-End Multi-Resolution MIL}
UFC-MIL $f_{\theta}=\{f_{\theta_r}\}_{r=1:R}$ consists of resolution-wise models, where $r=1$ is the lowest resolution, $r=2$ is the next highest, and $r=R$ represents the finest resolution index available, respectively. We denote the individual sample index of the dataset $\mathcal{D}$ as $i$. Given extracted features $Z_i^r=[z_{(i,1)}^r,\cdots,z^r_{(i,n_r)}]$ from $r$, where $n_r$ is the number of patches, $f_{\theta_r}$ produces $\hat{p}_i^r\in\mathbb{R}^{1\times{C}}$ as the aggregated prediction, and $\mathbf{\hat{p}}^r_i=[\hat{p}^r_{(i,1)},\cdots,\hat{p}^r_{(i,n_r)}]\in\mathbb{R}^{n_r\times{C}}$ for individual instance predictions. Here, $C$ is binary, and all $\hat{p}$ are softmax-probabilized vectors. $\hat{p}^r_i$ is trained by cross-entropy, which uses the given weak label $Y_i$:
\begin{equation}
    \mathcal{L}^r_{i,CE}=-\sum_{c\in \{0,1\}}(1-Y_i)\log{\hat{p}_{i}^r[c]}
\end{equation}
where $[c]$ denotes the $c$-th dimension of the vector. Additionally, we propose a patch-wise (PW) loss to correct per-instance predictions, strictly maintaining MIL's core assumptions:
\begin{multline}\label{eq:PW}
    \mathcal{L}_{i,PW}^{r}=(1-Y_i)\times\frac{1}{n_r}\underbrace{\sum_{n=1}^{n_r}\text{ReLU}(\hat{p}^r_{(i,n)}[1]-\delta)}_{\text{Negative: Without Exception}}\space+\\Y_i\times\underbrace{\text{ReLU}\left(-\max(\mathbf{\hat{p}}^{r}_{i}[1])+(1-\delta)\right)}_{\text{Positive: At Least One}}\space{.}
\end{multline}
This term tackles two issues: the non-differentiability of the $\operatorname{argmax}$ operation and unknown labels for individual instances.
Although $\mathbf{\hat{p}}^r_i$ can be trained separately using $\operatorname{argmax}$ when the label is $Y_i\geq1$, this non-differentiable operation impedes individual instance analysis~\cite{li2021localization}. We avoid this problem by directly regularizing continuous prediction probabilities. Negative samples are constrained to have all instances match the ground-truth label, while positive samples require at least one instance to match. Furthermore, due to the nature of weak supervision, $\mathbf{\hat{p}}^r_i$ predictions might inherently contain uncertainty. Therefore, the loss of PW employs a $\text{ReLU}(\cdot)$ with margin
$\delta<0.5$, which can be the space for the decision on unknown labels, causing the model to incorporate a corresponding level of uncertainty.

UFC-MIL is trained jointly for all samples $i$ and resolutions $r$:
\begin{equation}\label{eq:total_loss}
    \mathcal{L}=\sum_i\sum_r{(\mathcal{L}_{i,CE}^r+\mathcal{L}_{i,PW}^r}).
\end{equation}

\subsection{UFC-MIL Architecture}
\subsubsection{Efficient Attention Block}
Recent self-attention-based MILs have shown impressive performance, highlighting the ability to identify key features in instances~\cite{xiong2023diagnose,chen2022scaling,huang2023cross}. However, the computational complexity $\mathcal{O}(N^2)$ is prohibitive for the WSI analysis, which has numerous instances. Motivated by \cite{shao2021transmil}, we leverage a Nyström-based method~\cite{xiong2021nystromformer} to ease it. The input $X^r_i$, concatenated with a learnable class token $cls_i^{r}\in\mathbb{R}^{1\times{d}}$, is fed into the attention block, producing an output $\tilde{Z}^r_i\in\mathbb{R}^{(n_r+1)\times{d}}$.

\subsubsection{Topological Neighbor Attention Module}
MIL, which analyzes thousands of patches, enables structural approaches using position information~\cite{shao2021transmil}. However, an instance bag is an unordered collection of patches, rendering absolute positional information ambiguous. Instead, for spatially invariant patches, their relative contiguity is more significant~\cite{kang2023benchmarking}. Thus, we propose a topological neighbor attention module (TNAM) to aggregate patch spatial information. Given the adjacency matrix $\mathcal{A}^r_i\in\mathbb{R}^{n_r\times{n_r}}$ for patches, we define $\mathcal{N}^r_{(i,n)}$ as the set of neighbors adjacent to patch $x^r_{(i,n)}$.
The attention score $s^r_{(i,n)}$ contributed by $\mathcal{N}^r_{(i,n)}$ to the instance is:
\begin{equation}\label{eq:a_n}
    s^r_{(i,n)}=\frac{e^{\{{w}^T(\text{tanh}(\textit{\textbf{A}}_{t}{\tilde{z}}^r_{(i,n)})\,{\odot}\,\sigma(\textit{\textbf{A}}_{s}{\tilde{z}}^r_{(i,n)}))\}}}
    {\sum_{k\in\mathcal{N}^r_{(i,n)}}{e^{\{{w}^T(\text{tanh}(\textit{\textbf{A}}_{t}\tilde{z}^r_{(i,k)})\,{\odot}\,\sigma(\textit{\textbf{A}}_{s}\tilde{z}^r_{(i,k)}))\}}}}
\end{equation}
where $w\in\mathbb{R}^d$ and $\textbf{A}_{t,s}\in\mathbb{R}^{d\times{d}}$ are learnable parameters while $\sigma(\cdot)$ indicates sigmoid. The aggregated neighbor information for each instance is as follows:
\begin{equation}
    t^r_{(i,n)}=\sum_{k\in\mathcal{N}^r_{(i,n)}}s^r_{(i,k)}{\times}\tilde{z}^r_{(i,k)}\in\mathbb{R}^{d}.
\end{equation}
The resulting matrix $T^r_i=[t^r_{(i,1)},\cdots,t^r_{(i,n_r)}]\in\mathbb{R}^{{n_r}\times{d}}$ is combined with $\tilde{Z}^r_{i}$ using a residual sum. We specifically exclude $cls_i^{r}$ during TNAM and the residual summation, and then add it back afterward.
\subsubsection{Uncertainty-Masked Cross-Attention}
We propose uncertainty-masked cross-attention to emulate the expert's pattern of focusing more on uncertain areas and observing them with a magnified view. Since the proposed PW loss enables predictions for all patches, it allows us to quantify their uncertainty like human experts. For all $r\geq{1}$, patch-wise entropy at resolution $r$ is given by Equ.~\ref{eq:entropy}:
\begin{equation}\label{eq:entropy}
    H_i^{r}=-\sum_{c\in\{0,1\}}(\mathbf{\hat{p}}^{r}_i[c]\times\log_2{\mathbf{\hat{p}}^{r}_i[c]})\in\mathbb{R}^{n_{r}}.
\end{equation}
High entropy identifies patches that need focus, but their conversion to binary indicators prevents differentiation~\cite{li2021localization}. Instead of making each $f_{\theta_r}$ a sub-optimal~\cite{xiong2023diagnose}, we utilize Gumbel-softmax~\cite{jang2016categorical} to create a differentiable binary mask $\mathbf{m}^{r}_i=[m^{r}_{(i,1)},\cdots,m^{r}_{(i,n_{r})}]\in\mathbb{R}^{n_{r}}$:
\begin{multline}
    {m}^{r}_{(i,n)}=
    \\
    \mathbb{1}\left(\frac{e^{\{(\log(H^{r}_{i}[n])+g)/\tau\}}}{\sum_{c\in\{0,1\}}e^{\{(\log(1-c+(-1)^{1-c}H_{i}^{r}[n])+g)/\tau\}}}>0.5\right)
    \\
    \text{, where }\tau=1\text{ and }g\sim\text{Gumbel}(0,0.2).
\end{multline}
Using $\mathbf{m}^{r}_i$, we fuse features from resolution $r$ with those from $r+1$. After detaching $cls_i^{r+1}$ from $\tilde{Z}_i^{r+1}$, we create the features to be cross-attented as $(1-\text{Repeat(}\mathbf{m}^r_i,n_{r+1}/n_r)){\odot}\tilde{Z}^{r+1}_i+\text{Repeat}(\mathbf{m}^r_i\odot{Z}_i^r,n_{r+1}/n_r)$. 
Here, the function $\text{Repeat}(\mathbf{v}, j)$ duplicates and stacks a tensor, \textit{e}.\textit{g}., $\text{Repeat}([v_1,v_2],j)=[v_{(1,1)},$$\cdots,v_{(1,j)},v_{(2,1)},\cdots,v_{(2,j)}]$. The $cls^{r+1}_i$ token is then re-concatenated, and this combined feature vector, now of size in $\mathbb{R}^{n_{r+1}+1}$, is used in a cross-attention operation with $\tilde{Z}^{r+1}_i$.

\subsubsection{Identical Dimension Reduction Network}
The aggregated predictions from $\hat{p}_i^r$ and the multiple predictions from $\mathbf{\hat{p}}^{r}_i$ at each resolution $r$ are handled by an identical dimension reduction network. It consists of two linear layers with $\text{ELU}$ activation~\cite{clevert2015fast} followed by a $p=0.5$ dropout layer in between, which outputs $C$-dimensional probability.


\subsection{Sample and Resolution-Wise Label Smoothing}
We introduce an inference-free model calibration that leverages UFC-MIL's prediction on multiple patches. Inspired by \cite{park2025uncertainty}, we employ sample-wise label smoothing while simultaneously suggesting considering the heterogeneity across resolutions of MRMIL. Therefore, we propose a sample- and resolution-wise label smoothing (SRLS).

For all samples $i$ and resolutions $r$ the mean and standard deviation of $\mathbf{\hat{p}}^r_i$ are recorded as $\mathcal{M}^r\leftarrow\bigcup_{i\in\mathcal{D}}\text{mean}\left(H(\mathbf{\hat{p}}^r_i)\right)$ and $\mathcal{S}^r\leftarrow\bigcup_{i\in\mathcal{D}}\text{std}\left(H(\mathbf{\hat{p}}^r_i)\right)$ at the pre-defined training epoch. Then, each sets are min-max scaled as $\tilde{\mathcal{M}}_i^r$ and $\tilde{\mathcal{S}}^r_i$.
The label smoothing factor is defined as
\begin{equation}
    \varepsilon^r_i=\frac{1}{2}(\tilde{\mathcal{M}}^r_i+\tilde{\mathcal{S}}^r_i)\times\alpha
\end{equation}
where $\alpha$ is the temperature scaling factor.
For the sample $i$ and resolution $r$, the smoothed label is given as follows:
\begin{equation}
    \tilde{Y}^r_i=(1-\varepsilon^r_i)Y_i+\varepsilon^r_i/C.
\end{equation}
We conduct additional calibration training using the soft label $\tilde{Y}^r_i$ and only the $\mathcal{L}^r_{i,CE}$ term for the last several epochs, which is detailed in the pseudo-algorithm in the supplementary material. Since UFC-MIL measures patch-wise entropy for individual instances, it can perform calibration training directly without additional inference steps. For clarity, we would refer to UFC-MIL that has undergone SRLS calibration training as UFC-MIL$^\bigstar$.
\begin{table*}[t]
\centering
\resizebox{\textwidth}{!}{%
\begin{tabular}{cl|cccc|cccc|cccc}
\hline
                                                                                                                                  & \multicolumn{1}{c|}{}                        & \multicolumn{4}{c|}{CAMELYON16~\cite{bejnordi2017diagnostic}}                                                                                                                                                                                                                                     & \multicolumn{4}{c|}{DHMC~\cite{wei2019pathologist}}                                                                                                                                                                                                                                                   & \multicolumn{4}{c}{{\color[HTML]{000000} BCNB~\cite{xu2021predicting}}}                                                                                                                                                                                                                                                                                                                       \\ \cline{3-14} 
\multirow{-2}{*}{\begin{tabular}[c]{@{}c@{}}Calibration\\ Method\end{tabular}}                                                    & \multicolumn{1}{c|}{\multirow{-2}{*}{MRMIL}} & ECE $\downarrow$                                                         & R@10\% $\uparrow$                                                    & R@30\% $\uparrow$                                                    & Accuracy $\uparrow$                                                      & ECE $\downarrow$                                                         & R@10\% $\uparrow$                                                    & R@30\% $\uparrow$                                                        & Accuracy $\uparrow$                                                      & {\color[HTML]{000000} ECE $\downarrow$}                                                         & {\color[HTML]{000000} R@10\% $\uparrow$}                                                    & {\color[HTML]{000000} R@30\% $\uparrow$}                                                    & {\color[HTML]{000000} Accuracy $\uparrow$}                                                      \\ \hline
                                                                                                                                  & DS-MIL~\cite{li2021dual}                     & \begin{tabular}[c]{@{}c@{}}0.086\\ \small{(0.002)}\end{tabular}          & \textbf{\begin{tabular}[c]{@{}c@{}}1.0\\ \small{(0.0)}\end{tabular}} & \begin{tabular}[c]{@{}c@{}}0.914\\ \small{(0.026)}\end{tabular}      & \begin{tabular}[c]{@{}c@{}}0.909\\ \small{(0.011)}\end{tabular}          & \begin{tabular}[c]{@{}c@{}}0.236\\ \small{(0.024)}\end{tabular}          & \begin{tabular}[c]{@{}c@{}}0.851\\ \small{(0.135)}\end{tabular}      & \begin{tabular}[c]{@{}c@{}}0.839\\ \small{(0.069)}\end{tabular}          & \begin{tabular}[c]{@{}c@{}}0.751\\ \small{(0.027)}\end{tabular}          & {\color[HTML]{000000} \begin{tabular}[c]{@{}c@{}}0.214\\ \small{(0.029)}\end{tabular}}          & {\color[HTML]{000000} \begin{tabular}[c]{@{}c@{}}0.980\\ \small{(0.034)}\end{tabular}}      & {\color[HTML]{000000} \begin{tabular}[c]{@{}c@{}}0.968\\ \small{(0.040)}\end{tabular}}      & {\color[HTML]{000000} \begin{tabular}[c]{@{}c@{}}0.767\\ \small{(0.033)}\end{tabular}}          \\
                                                                                                                                  & HAG-MIL~\cite{xiong2023diagnose}             & \begin{tabular}[c]{@{}c@{}}0.147\\ \small{(0.016})\end{tabular}          & \textbf{\begin{tabular}[c]{@{}c@{}}1.0\\ \small{(0.0)}\end{tabular}} & \begin{tabular}[c]{@{}c@{}}0.969\\ \small{(0.031)}\end{tabular}      & \begin{tabular}[c]{@{}c@{}}0.847\\ \small{(0.011)}\end{tabular}          & \begin{tabular}[c]{@{}c@{}}0.243\\ \small{(0.021)}\end{tabular}          & \begin{tabular}[c]{@{}c@{}}0.667\\ \small{(0.471)}\end{tabular}      & \begin{tabular}[c]{@{}c@{}}0.747\\ \small{(0.188)}\end{tabular}          & \begin{tabular}[c]{@{}c@{}}0.753\\ \small{(0.032)}\end{tabular}          & {\color[HTML]{000000} \begin{tabular}[c]{@{}c@{}}0.186\\ \small{(0.006)}\end{tabular}}          & {\color[HTML]{000000} \begin{tabular}[c]{@{}c@{}}0.963\\ \small{(0.064)}\end{tabular}}      & {\color[HTML]{000000} \begin{tabular}[c]{@{}c@{}}0.980\\ \small{(0.034)}\end{tabular}}      & {\color[HTML]{000000} \begin{tabular}[c]{@{}c@{}}0.805\\ \small{(0.003)}\end{tabular}}          \\
                                                                                                                                  & Godson et al.~\cite{godson2023multi}         & \begin{tabular}[c]{@{}c@{}}0.074\\ \small{(0.018)}\end{tabular}          & \textbf{\begin{tabular}[c]{@{}c@{}}1.0\\ \small{(0.0)}\end{tabular}} & \begin{tabular}[c]{@{}c@{}}0.977\\ \small{(0.040)}\end{tabular}      & \begin{tabular}[c]{@{}c@{}}0.894\\ \small{(0.020)}\end{tabular}          & \begin{tabular}[c]{@{}c@{}}0.224\\ \small{(0.013)}\end{tabular}          & \begin{tabular}[c]{@{}c@{}}0.966\\ \small{(0.066)}\end{tabular}      & \begin{tabular}[c]{@{}c@{}}0.857\\ \small{(0.038)}\end{tabular}          & \begin{tabular}[c]{@{}c@{}}0.758\\ \small{(0.022)}\end{tabular}          & {\color[HTML]{000000} \begin{tabular}[c]{@{}c@{}}0.186\\ \small{(0.017)}\end{tabular}}          & {\color[HTML]{000000} \textbf{\begin{tabular}[c]{@{}c@{}}1.0\\ \small{(0.0)}\end{tabular}}} & {\color[HTML]{000000} \begin{tabular}[c]{@{}c@{}}0.988\\ \small{(0.019)}\end{tabular}}      & {\color[HTML]{000000} \begin{tabular}[c]{@{}c@{}}0.802\\ \small{(0.014)}\end{tabular}}          \\
\multirow{-7}{*}{-}                                                                                                               & UFC-MIL                                      & \begin{tabular}[c]{@{}c@{}}0.086\\ \small{(0.037)}\end{tabular}          & \textbf{\begin{tabular}[c]{@{}c@{}}1.0\\ \small{(0.0)}\end{tabular}} & \textbf{\begin{tabular}[c]{@{}c@{}}1.0\\ \small{(0.0)}\end{tabular}} & \begin{tabular}[c]{@{}c@{}}0.917\\ \small{(0.038)}\end{tabular}          & \begin{tabular}[c]{@{}c@{}}0.202\\ \small{(0.017)}\end{tabular}          & \begin{tabular}[c]{@{}c@{}}0.986\\ \small{(0.045)}\end{tabular}      & \begin{tabular}[c]{@{}c@{}}0.891\\ \small{(0.061)}\end{tabular}          & \begin{tabular}[c]{@{}c@{}}0.793\\ \small{(0.026)}\end{tabular}          & {\color[HTML]{000000} \begin{tabular}[c]{@{}c@{}}0.112\\ \small{(0.019)}\end{tabular}}          & {\color[HTML]{000000} \textbf{\begin{tabular}[c]{@{}c@{}}1.0\\ \small{(0.0)}\end{tabular}}} & {\color[HTML]{000000} \begin{tabular}[c]{@{}c@{}}0.993\\ \small{(0.008)}\end{tabular}}      & {\color[HTML]{000000} \begin{tabular}[c]{@{}c@{}}0.804\\ \small{(0.018)}\end{tabular}}          \\
\rowcolor[HTML]{EFEFEF} 
\cellcolor[HTML]{EFEFEF}                                                                                                          & DS-MIL~\cite{li2021dual}                     & \begin{tabular}[c]{@{}c@{}}0.069\\ \small{(0.011)}\end{tabular}          & \textbf{\begin{tabular}[c]{@{}c@{}}1.0\\ \small{(0.0)}\end{tabular}} & \textbf{\begin{tabular}[c]{@{}c@{}}1.0\\ \small{(0.0)}\end{tabular}} & \begin{tabular}[c]{@{}c@{}}0.925\\ \small{(0.004)}\end{tabular}          & \begin{tabular}[c]{@{}c@{}}0.226\\ \small{(0.029)}\end{tabular}          & \begin{tabular}[c]{@{}c@{}}0.916\\ \small{(0.117)}\end{tabular}      & \begin{tabular}[c]{@{}c@{}}0.846\\ \small{(0.059)}\end{tabular}          & \begin{tabular}[c]{@{}c@{}}0.758\\ \small{(0.016)}\end{tabular}          & {\color[HTML]{000000} \begin{tabular}[c]{@{}c@{}}0.188\\ \small{(0.018)}\end{tabular}}          & {\color[HTML]{000000} \begin{tabular}[c]{@{}c@{}}0.961\\ \small{(0.033)}\end{tabular}}      & {\color[HTML]{000000} \begin{tabular}[c]{@{}c@{}}0.988\\ \small{(0.010)}\end{tabular}}      & {\color[HTML]{000000} \begin{tabular}[c]{@{}c@{}}0.801\\ \small{(0.007)}\end{tabular}}          \\
\rowcolor[HTML]{EFEFEF} 
\cellcolor[HTML]{EFEFEF}                                                                                                          & HAG-MIL~\cite{xiong2023diagnose}             & \begin{tabular}[c]{@{}c@{}}0.146\\ \small{(0.074)}\end{tabular}          & \textbf{\begin{tabular}[c]{@{}c@{}}1.0\\ \small{(0.0)}\end{tabular}} & \begin{tabular}[c]{@{}c@{}}0.988\\ \small{(0.019)}\end{tabular}      & \begin{tabular}[c]{@{}c@{}}0.847\\ \small{(0.022)}\end{tabular}          & \begin{tabular}[c]{@{}c@{}}0.232\\ \small{(0.027)}\end{tabular}          & \begin{tabular}[c]{@{}c@{}}0.4\\ \small{(0.547)}\end{tabular}        & \begin{tabular}[c]{@{}c@{}}0.772\\ \small{(0.178)}\end{tabular}          & \begin{tabular}[c]{@{}c@{}}0.758\\ \small{(0.036)}\end{tabular}          & {\color[HTML]{000000} \begin{tabular}[c]{@{}c@{}}0.187\\ \small{(0.021)}\end{tabular}}          & {\color[HTML]{000000} \begin{tabular}[c]{@{}c@{}}0.965\\ \small{(0.060)}\end{tabular}}      & {\color[HTML]{000000} \begin{tabular}[c]{@{}c@{}}0.977\\ \small{(0.041)}\end{tabular}}      & {\color[HTML]{000000} \begin{tabular}[c]{@{}c@{}}0.810\\ \small{(0.024)}\end{tabular}}          \\
\rowcolor[HTML]{EFEFEF} 
\cellcolor[HTML]{EFEFEF}                                                                                                          & Godson et al.~\cite{godson2023multi}         & \begin{tabular}[c]{@{}c@{}}0.075\\ \small{(0.015)}\end{tabular}          & \textbf{\begin{tabular}[c]{@{}c@{}}1.0\\ \small{(0.0)}\end{tabular}} & \begin{tabular}[c]{@{}c@{}}0.985\\ \small{(0.036)}\end{tabular}      & \begin{tabular}[c]{@{}c@{}}0.901\\ \small{(0.018)}\end{tabular}          & \begin{tabular}[c]{@{}c@{}}0.206\\ \small{(0.021)}\end{tabular}          & \begin{tabular}[c]{@{}c@{}}0.983\\ \small{(0.052)}\end{tabular}      & \begin{tabular}[c]{@{}c@{}}0.894\\ \small{(0.065)}\end{tabular}          & \begin{tabular}[c]{@{}c@{}}0.771\\ \small{(0.025)}\end{tabular}          & {\color[HTML]{000000} \begin{tabular}[c]{@{}c@{}}0.175\\ \small{(0.007)}\end{tabular}}          & {\color[HTML]{000000} \textbf{\begin{tabular}[c]{@{}c@{}}1.0\\ \small{(0.0)}\end{tabular}}} & {\color[HTML]{000000} \begin{tabular}[c]{@{}c@{}}0.971\\ \small{(0.016)}\end{tabular}}      & {\color[HTML]{000000} \begin{tabular}[c]{@{}c@{}}0.816\\ \small{(0.005)}\end{tabular}}          \\
\rowcolor[HTML]{EFEFEF} 
\multirow{-7}{*}{\cellcolor[HTML]{EFEFEF}\begin{tabular}[c]{@{}c@{}}Temperature\\ Scaling~\cite{guo2017calibration}\end{tabular}} & UFC-MIL                                      & \begin{tabular}[c]{@{}c@{}}0.083\\ \small{(0.015)}\end{tabular}          & \textbf{\begin{tabular}[c]{@{}c@{}}1.0\\ \small{(0.0)}\end{tabular}} & \begin{tabular}[c]{@{}c@{}}0.987\\ \small{(0.022)}\end{tabular}      & \begin{tabular}[c]{@{}c@{}}0.917\\ \small{(0.018)}\end{tabular}          & \begin{tabular}[c]{@{}c@{}}0.203\\ \small{(0.027)}\end{tabular}          & \begin{tabular}[c]{@{}c@{}}0.893\\ \small{(0.125)}\end{tabular}      & \begin{tabular}[c]{@{}c@{}}0.882\\ \small{(0.027)}\end{tabular}          & \begin{tabular}[c]{@{}c@{}}0.794\\ \small{(0.034)}\end{tabular}          & {\color[HTML]{000000} \begin{tabular}[c]{@{}c@{}}0.107\\ \small{(0.019)}\end{tabular}}          & {\color[HTML]{000000} \begin{tabular}[c]{@{}c@{}}0.982\\ \small{(0.030)}\end{tabular}}      & {\color[HTML]{000000} \begin{tabular}[c]{@{}c@{}}0.977\\ \small{(0.019)}\end{tabular}}      & {\color[HTML]{000000} \begin{tabular}[c]{@{}c@{}}0.812\\ \small{(0.018)}\end{tabular}}          \\
                                                                                                                                  & DS-MIL~\cite{li2021dual}                     & \begin{tabular}[c]{@{}c@{}}0.077\\ \small{(0.026)}\end{tabular}          & \begin{tabular}[c]{@{}c@{}}0.970\\ \small{(0.052)}\end{tabular}      & \begin{tabular}[c]{@{}c@{}}0.939\\ \small{(0.105)}\end{tabular}      & \begin{tabular}[c]{@{}c@{}}0.927\\ \small{(0.019)}\end{tabular}          & \begin{tabular}[c]{@{}c@{}}0.202\\ \small{(0.030)}\end{tabular}          & \begin{tabular}[c]{@{}c@{}}0.866\\ \small{(0.097)}\end{tabular}      & \begin{tabular}[c]{@{}c@{}}0.877\\ \small{(0.066)}\end{tabular}          & \begin{tabular}[c]{@{}c@{}}0.736\\ \small{(0.035)}\end{tabular}          & {\color[HTML]{000000} \begin{tabular}[c]{@{}c@{}}0.116\\ \small{(0.021)}\end{tabular}}          & {\color[HTML]{000000} \begin{tabular}[c]{@{}c@{}}0.980\\ \small{(0.034)}\end{tabular}}      & {\color[HTML]{000000} \begin{tabular}[c]{@{}c@{}}0.951\\ \small{(0.028)}\end{tabular}}      & {\color[HTML]{000000} \begin{tabular}[c]{@{}c@{}}0.775\\ \small{(0.029)}\end{tabular}}          \\
                                                                                                                                  & HAG-MIL~\cite{xiong2023diagnose}             & \begin{tabular}[c]{@{}c@{}}0.093\\ \small{(0.014)}\end{tabular}          & \textbf{\begin{tabular}[c]{@{}c@{}}1.0\\ \small{(0.0)}\end{tabular}} & \begin{tabular}[c]{@{}c@{}}0.937\\ \small{(0.012)}\end{tabular}      & \begin{tabular}[c]{@{}c@{}}0.870\\ \small{(0.031)}\end{tabular}          & \begin{tabular}[c]{@{}c@{}}0.217\\ \small{(0.017)}\end{tabular}          & \begin{tabular}[c]{@{}c@{}}0.913\\ \small{(0.093)}\end{tabular}      & \begin{tabular}[c]{@{}c@{}}0.816\\ \small{(0.078)}\end{tabular}          & \begin{tabular}[c]{@{}c@{}}0.751\\ \small{(0.026)}\end{tabular}          & {\color[HTML]{000000} \begin{tabular}[c]{@{}c@{}}0.150\\ \small{(0.007)}\end{tabular}}          & {\color[HTML]{000000} \begin{tabular}[c]{@{}c@{}}0.955\\ \small{(0.031)}\end{tabular}}      & {\color[HTML]{000000} \begin{tabular}[c]{@{}c@{}}0.958\\ \small{(0.025)}\end{tabular}}      & {\color[HTML]{000000} \begin{tabular}[c]{@{}c@{}}0.808\\ \small{(0.005)}\end{tabular}}          \\
                                                                                                                                  & Godson et al.~\cite{godson2023multi}         & \begin{tabular}[c]{@{}c@{}}0.070\\ \small{(0.019)}\end{tabular}          & \textbf{\begin{tabular}[c]{@{}c@{}}1.0\\ \small{(0.0)}\end{tabular}} & \begin{tabular}[c]{@{}c@{}}0.976\\ \small{(0.047)}\end{tabular}      & \begin{tabular}[c]{@{}c@{}}0.895\\ \small{(0.021)}\end{tabular}          & \begin{tabular}[c]{@{}c@{}}0.211\\ \small{(0.006)}\end{tabular}          & \begin{tabular}[c]{@{}c@{}}0.960\\ \small{(0.080)}\end{tabular}      & \begin{tabular}[c]{@{}c@{}}0.851\\ \small{(0.026)}\end{tabular}          & \begin{tabular}[c]{@{}c@{}}0.762\\ \small{(0.022)}\end{tabular}          & {\color[HTML]{000000} \begin{tabular}[c]{@{}c@{}}0.131\\ \small{(0.032)}\end{tabular}}          & {\color[HTML]{000000} \begin{tabular}[c]{@{}c@{}}0.968\\ \small{(0.029)}\end{tabular}}      & {\color[HTML]{000000} \begin{tabular}[c]{@{}c@{}}0.977\\ \small{(0.020)}\end{tabular}}      & {\color[HTML]{000000} \begin{tabular}[c]{@{}c@{}}0.804\\ \small{(0.024)}\end{tabular}}          \\
\multirow{-7}{*}{\begin{tabular}[c]{@{}c@{}}Label\\ Smoothing~\cite{szegedy2016rethinking}\end{tabular}}                          & UFC-MIL                                      & \begin{tabular}[c]{@{}c@{}}0.073\\ \small{(0.032)}\end{tabular}          & \textbf{\begin{tabular}[c]{@{}c@{}}1.0\\ \small{(0.0)}\end{tabular}} & \begin{tabular}[c]{@{}c@{}}0.996\\ \small{(0.012)}\end{tabular}      & \begin{tabular}[c]{@{}c@{}}0.924\\ \small{(0.034)}\end{tabular}          & \begin{tabular}[c]{@{}c@{}}0.195\\ \small{(0.012)}\end{tabular}          & \begin{tabular}[c]{@{}c@{}}0.971\\ \small{(0.057)}\end{tabular}      & \begin{tabular}[c]{@{}c@{}}0.918\\ \small{(0.043)}\end{tabular}          & \begin{tabular}[c]{@{}c@{}}0.800\\ \small{(0.015)}\end{tabular}          & {\color[HTML]{000000} \begin{tabular}[c]{@{}c@{}}0.108\\ \small{(0.02)}\end{tabular}}           & {\color[HTML]{000000} \textbf{\begin{tabular}[c]{@{}c@{}}1.0\\ \small{(0.0)}\end{tabular}}} & {\color[HTML]{000000} \begin{tabular}[c]{@{}c@{}}0.983\\ \small{(0.016)}\end{tabular}}      & {\color[HTML]{000000} \begin{tabular}[c]{@{}c@{}}0.805\\ \small{(0.016)}\end{tabular}}          \\
\rowcolor[HTML]{EFEFEF} 
\cellcolor[HTML]{EFEFEF}                                                                                                          & DS-MIL~\cite{li2021dual}                     & \begin{tabular}[c]{@{}c@{}}0.061\\ \small{(0.009)}\end{tabular}          & \textbf{\begin{tabular}[c]{@{}c@{}}1.0\\ \small{(0.0)}\end{tabular}} & \begin{tabular}[c]{@{}c@{}}0.966\\ \small{(0.047)}\end{tabular}      & \begin{tabular}[c]{@{}c@{}}0.930\\ \small{(0.001)}\end{tabular}          & \begin{tabular}[c]{@{}c@{}}0.219\\ \small{(0.008)}\end{tabular}          & \begin{tabular}[c]{@{}c@{}}0.866\\ \small{(0.141)}\end{tabular}      & \begin{tabular}[c]{@{}c@{}}0.831\\ \small{(0.098)}\end{tabular}          & \begin{tabular}[c]{@{}c@{}}0.768\\ \small{(0.027)}\end{tabular}          & {\color[HTML]{000000} \begin{tabular}[c]{@{}c@{}}0.189\\ \small{(0.022)}\end{tabular}}          & {\color[HTML]{000000} \begin{tabular}[c]{@{}c@{}}0.958\\ \small{(0.072)}\end{tabular}}      & {\color[HTML]{000000} \begin{tabular}[c]{@{}c@{}}0.974\\ \small{(0.043)}\end{tabular}}      & {\color[HTML]{000000} \begin{tabular}[c]{@{}c@{}}0.761\\ \small{(0.041)}\end{tabular}}          \\
\rowcolor[HTML]{EFEFEF} 
\cellcolor[HTML]{EFEFEF}                                                                                                          & HAG-MIL~\cite{xiong2023diagnose}             & \begin{tabular}[c]{@{}c@{}}0.119\\ \small{(0.054)}\end{tabular}          & \begin{tabular}[c]{@{}c@{}}0.952\\ \small{(0.082)}\end{tabular}      & \begin{tabular}[c]{@{}c@{}}0.932\\ \small{(0.023)}\end{tabular}      & \begin{tabular}[c]{@{}c@{}}0.865\\ \small{(0.031)}\end{tabular}          & \begin{tabular}[c]{@{}c@{}}0.206\\ \small{(0.018)}\end{tabular}          & \begin{tabular}[c]{@{}c@{}}0.931\\ \small{(0.112)}\end{tabular}      & \begin{tabular}[c]{@{}c@{}}0.802\\ \small{(0.083)}\end{tabular}          & \begin{tabular}[c]{@{}c@{}}0.743\\ \small{(0.034)}\end{tabular}          & {\color[HTML]{000000} \begin{tabular}[c]{@{}c@{}}0.178\\ \small{(0.014)}\end{tabular}}          & {\color[HTML]{000000} \begin{tabular}[c]{@{}c@{}}0.963\\ \small{(0.018)}\end{tabular}}      & {\color[HTML]{000000} \begin{tabular}[c]{@{}c@{}}0.974\\ \small{(0.012)}\end{tabular}}      & {\color[HTML]{000000} \begin{tabular}[c]{@{}c@{}}0.794\\ \small{(0.026)}\end{tabular}}          \\
\rowcolor[HTML]{EFEFEF} 
\cellcolor[HTML]{EFEFEF}                                                                                                          & Godson et al.~\cite{godson2023multi}         & \begin{tabular}[c]{@{}c@{}}0.0732\\ \small{(0.015)}\end{tabular}         & \textbf{\begin{tabular}[c]{@{}c@{}}1.0\\ \small{(0.0)}\end{tabular}} & \begin{tabular}[c]{@{}c@{}}0.967\\ \small{(0.050)}\end{tabular}      & \begin{tabular}[c]{@{}c@{}}0.903\\ \small{(0.021)}\end{tabular}          & \begin{tabular}[c]{@{}c@{}}0.205\\ \small{(0.016)}\end{tabular}          & \begin{tabular}[c]{@{}c@{}}0.744\\ \small{(0.309)}\end{tabular}      & \begin{tabular}[c]{@{}c@{}}0.848\\ \small{(0.051)}\end{tabular}          & \begin{tabular}[c]{@{}c@{}}0.762\\ \small{90.024)}\end{tabular}          & {\color[HTML]{000000} \begin{tabular}[c]{@{}c@{}}0.170\\ \small{(0.020)}\end{tabular}}          & {\color[HTML]{000000} \textbf{\begin{tabular}[c]{@{}c@{}}1.0\\ \small{(0.0)}\end{tabular}}} & {\color[HTML]{000000} \begin{tabular}[c]{@{}c@{}}0.981\\ \small{(0.013)}\end{tabular}}      & {\color[HTML]{000000} \begin{tabular}[c]{@{}c@{}}0.792\\ \small{(0.021)}\end{tabular}}          \\
\rowcolor[HTML]{EFEFEF} 
\multirow{-7}{*}{\cellcolor[HTML]{EFEFEF}\begin{tabular}[c]{@{}c@{}}M.C.\\ Dropout$^\dagger$~\cite{gal2016dropout}\end{tabular}}  & UFC-MIL                                      & \begin{tabular}[c]{@{}c@{}}0.088\\ \small{(0.039)}\end{tabular}          & \textbf{\begin{tabular}[c]{@{}c@{}}1.0\\ \small{(0.0)}\end{tabular}} & \textbf{\begin{tabular}[c]{@{}c@{}}1.0\\ \small{(0.0)}\end{tabular}} & \begin{tabular}[c]{@{}c@{}}0.902\\ \small{(0.055)}\end{tabular}          & \begin{tabular}[c]{@{}c@{}}0.197\\ \small{(0.018)}\end{tabular}          & \begin{tabular}[c]{@{}c@{}}0.955\\ \small{(0.095)}\end{tabular}      & \begin{tabular}[c]{@{}c@{}}0.923\\ \small{(0.074)}\end{tabular}          & \begin{tabular}[c]{@{}c@{}}0.804\\ \small{(0.023)}\end{tabular}          & {\color[HTML]{000000} \begin{tabular}[c]{@{}c@{}}0.108\\ \small{(0.020)}\end{tabular}}          & {\color[HTML]{000000} \begin{tabular}[c]{@{}c@{}}0.994\\ \small{(0.016)}\end{tabular}}      & {\color[HTML]{000000} \begin{tabular}[c]{@{}c@{}}0.998\\ \small{(0.005)}\end{tabular}}      & {\color[HTML]{000000} \begin{tabular}[c]{@{}c@{}}0.803\\ \small{(0.018)}\end{tabular}}          \\
                                                                                                                                  & DS-MIL~\cite{li2021dual}                     & 0.072                                                                    & \textbf{1.0}                                                         & 0.954                                                                & 0.930                                                                    & 0.212                                                                    & \textbf{1.0}                                                         & 0.888                                                                    & 0.755                                                                    & {\color[HTML]{000000} 0.122}                                                                    & {\color[HTML]{000000} \textbf{1.0}}                                                         & {\color[HTML]{000000} 0.989}                                                                & {\color[HTML]{000000} 0.770}                                                                    \\
                                                                                                                                  & HAG-MIL~\cite{xiong2023diagnose}             & 0.100                                                                    & \textbf{1.0}                                                         & 0.962                                                                & 0.875                                                                    & 0.239                                                                    & \textbf{1.0}                                                         & 0.8                                                                      & 0.755                                                                    & {\color[HTML]{000000} 0.123}                                                                    & {\color[HTML]{000000} \textbf{1.0}}                                                         & {\color[HTML]{000000} 0.989}                                                                & {\color[HTML]{000000} 0.808}                                                                    \\
                                                                                                                                  & Godson et al.~\cite{godson2023multi}         & 0.078                                                                    & \textbf{1.0}                                                         & \textbf{1.0}                                                         & 0.899                                                                    & 0.256                                                                    & \textbf{1.0}                                                         & 0.833                                                                    & 0.773                                                                    & {\color[HTML]{000000} 0.144}                                                                    & {\color[HTML]{000000} \textbf{1.0}}                                                         & {\color[HTML]{000000} 0.978}                                                                & {\color[HTML]{000000} 0.794}                                                                    \\
\multirow{-4}{*}{\begin{tabular}[c]{@{}c@{}}Deep\\ Ensembles$^\dagger$~\cite{lakshminarayanan2017simple}\end{tabular}}            & UFC-MIL                                      & 0.062                                                                    & \textbf{1.0}                                                         & \textbf{1.0}                                                         & 0.930                                                                    & 0.212                                                                    & \textbf{1.0}                                                         & 0.875                                                                    & 0.811                                                                    & {\color[HTML]{000000} 0.130}                                                                    & {\color[HTML]{000000} \textbf{1.0}}                                                         & {\color[HTML]{000000} 0.989}                                                                & {\color[HTML]{000000} 0.818}                                                                    \\
\rowcolor[HTML]{EFEFEF} 
\cellcolor[HTML]{EFEFEF}                                                                                                          & DS-MIL~\cite{li2021dual}                     & \begin{tabular}[c]{@{}c@{}}0.102\\ \small{(0.034)}\end{tabular}          & \textbf{\begin{tabular}[c]{@{}c@{}}1.0\\ \small{(0.0)}\end{tabular}} & \begin{tabular}[c]{@{}c@{}}0.986\\ \small{(0.023)}\end{tabular}      & \begin{tabular}[c]{@{}c@{}}0.894\\ \small{(0.031)}\end{tabular}          & \begin{tabular}[c]{@{}c@{}}0.240\\ \small{(0.038)}\end{tabular}          & \begin{tabular}[c]{@{}c@{}}0.778\\ \small{(0.154)}\end{tabular}      & \begin{tabular}[c]{@{}c@{}}0.843\\ \small{(0.061)}\end{tabular}          & \begin{tabular}[c]{@{}c@{}}0.743\\ \small{(0.048)}\end{tabular}          & {\color[HTML]{000000} \begin{tabular}[c]{@{}c@{}}0.153\\ \small{(0.003)}\end{tabular}}          & {\color[HTML]{000000} \begin{tabular}[c]{@{}c@{}}0.963\\ \small{(0.031)}\end{tabular}}      & {\color[HTML]{000000} \begin{tabular}[c]{@{}c@{}}0.968\\ \small{(0.023)}\end{tabular}}      & {\color[HTML]{000000} \begin{tabular}[c]{@{}c@{}}0.781\\ \small{(0.023)}\end{tabular}}          \\
\rowcolor[HTML]{EFEFEF} 
\cellcolor[HTML]{EFEFEF}                                                                                                          & HAG-MIL~\cite{xiong2023diagnose}             & \begin{tabular}[c]{@{}c@{}}0.125\\ \small{(0.031)}\end{tabular}          & \textbf{\begin{tabular}[c]{@{}c@{}}1.0\\ \small{(0.0)}\end{tabular}} & \begin{tabular}[c]{@{}c@{}}0.965\\ \small{(0.041)}\end{tabular}      & \begin{tabular}[c]{@{}c@{}}0.837\\ \small{(0.037)}\end{tabular}          & \begin{tabular}[c]{@{}c@{}}0.223\\ \small{(0.025)}\end{tabular}          & \begin{tabular}[c]{@{}c@{}}0.946\\ \small{(0.086)}\end{tabular}      & \begin{tabular}[c]{@{}c@{}}0.842\\ \small{(0.091)}\end{tabular}          & \begin{tabular}[c]{@{}c@{}}0.755\\ \small{(0.028)}\end{tabular}          & {\color[HTML]{000000} \begin{tabular}[c]{@{}c@{}}0.145\\ \small{(0.005)}\end{tabular}}          & {\color[HTML]{000000} \begin{tabular}[c]{@{}c@{}}0.933\\ \small{(0.067)}\end{tabular}}      & {\color[HTML]{000000} \begin{tabular}[c]{@{}c@{}}0.974\\ \small{(0.023)}\end{tabular}}      & {\color[HTML]{000000} \begin{tabular}[c]{@{}c@{}}0.810\\ \small{(0.012)}\end{tabular}}          \\
\rowcolor[HTML]{EFEFEF} 
\cellcolor[HTML]{EFEFEF}                                                                                                          & Godson et al.~\cite{godson2023multi}         & \begin{tabular}[c]{@{}c@{}}0.062\\ \small{(0.016)}\end{tabular}          & \textbf{\begin{tabular}[c]{@{}c@{}}1.0\\ \small{(0.0)}\end{tabular}} & \begin{tabular}[c]{@{}c@{}}0.889\\ \small{(0.093)}\end{tabular}      & \begin{tabular}[c]{@{}c@{}}0.827\\ \small{(0.071)}\end{tabular}          & \begin{tabular}[c]{@{}c@{}}0.242\\ \small{(0.034)}\end{tabular}          & \begin{tabular}[c]{@{}c@{}}0.701\\ \small{(0.483)}\end{tabular}      & \begin{tabular}[c]{@{}c@{}}0.706\\ \small{(0.392)}\end{tabular}          & \begin{tabular}[c]{@{}c@{}}0.724\\ \small{(0.042)}\end{tabular}          & {\color[HTML]{000000} \begin{tabular}[c]{@{}c@{}}0.167\\ \small{(0.004)}\end{tabular}}          & {\color[HTML]{000000} \begin{tabular}[c]{@{}c@{}}0.929\\ \small{(0.043)}\end{tabular}}      & {\color[HTML]{000000} \begin{tabular}[c]{@{}c@{}}0.951\\ \small{(0.025)}\end{tabular}}      & {\color[HTML]{000000} \begin{tabular}[c]{@{}c@{}}0.800\\ \small{(0.009)}\end{tabular}}          \\
\rowcolor[HTML]{EFEFEF} 
\multirow{-7}{*}{\cellcolor[HTML]{EFEFEF}UDLS$^\dagger$~\cite{park2025uncertainty}}                                               & UFC-MIL                                      & \begin{tabular}[c]{@{}c@{}}0.112\\ \small{(0.045)}\end{tabular}          & \begin{tabular}[c]{@{}c@{}}0.958\\ \small{(0.072)}\end{tabular}      & \begin{tabular}[c]{@{}c@{}}0.883\\ \small{(0.013)}\end{tabular}      & \begin{tabular}[c]{@{}c@{}}0.868\\ \small{(0.031)}\end{tabular}          & \begin{tabular}[c]{@{}c@{}}0.214\\ \small{(0.026)}\end{tabular}          & \begin{tabular}[c]{@{}c@{}}0.983\\ \small{(0.052)}\end{tabular}      & \begin{tabular}[c]{@{}c@{}}0.908\\ \small{(0.034)}\end{tabular}          & \begin{tabular}[c]{@{}c@{}}0.783\\ \small{(0.023)}\end{tabular}          & {\color[HTML]{000000} \begin{tabular}[c]{@{}c@{}}0.086\\ \small{(0.028)}\end{tabular}}          & {\color[HTML]{000000} \begin{tabular}[c]{@{}c@{}}0.983\\ \small{(0.027)}\end{tabular}}      & {\color[HTML]{000000} \begin{tabular}[c]{@{}c@{}}0.983\\ \small{(0.022)}\end{tabular}}      & {\color[HTML]{000000} \begin{tabular}[c]{@{}c@{}}0.783\\ \small{(0.019)}\end{tabular}}          \\ \hline
\multicolumn{2}{c|}{UFC-MIL$^\bigstar$}                                                                                                                                          & \textbf{\begin{tabular}[c]{@{}c@{}}0.056\\ \small{(0.016)}\end{tabular}} & \textbf{\begin{tabular}[c]{@{}c@{}}1.0\\ \small{(0.0)}\end{tabular}} & \textbf{\begin{tabular}[c]{@{}c@{}}1.0\\ \small{(0.0)}\end{tabular}} & \textbf{\begin{tabular}[c]{@{}c@{}}0.941\\ \small{(0.011)}\end{tabular}} & \textbf{\begin{tabular}[c]{@{}c@{}}0.189\\ \small{(0.021)}\end{tabular}} & \textbf{\begin{tabular}[c]{@{}c@{}}1.0\\ \small{(0.0)}\end{tabular}} & \textbf{\begin{tabular}[c]{@{}c@{}}0.964\\ \small{(0.051)}\end{tabular}} & \textbf{\begin{tabular}[c]{@{}c@{}}0.812\\ \small{(0.021)}\end{tabular}} & {\color[HTML]{000000} \textbf{\begin{tabular}[c]{@{}c@{}}0.077\\ \small{(0.033)}\end{tabular}}} & {\color[HTML]{000000} \textbf{\begin{tabular}[c]{@{}c@{}}1.0\\ \small{(0.0)}\end{tabular}}} & {\color[HTML]{000000} \textbf{\begin{tabular}[c]{@{}c@{}}1.0\\ \small{(0.0)}\end{tabular}}} & {\color[HTML]{000000} \textbf{\begin{tabular}[c]{@{}c@{}}0.820\\ \small{(0.028)}\end{tabular}}} \\ \hline
\end{tabular}%
}
\caption{Quantitative results on CAMELYON16, DHMC, and BCNB datasets. We report the mean and standard deviation, with the latter indicated in parentheses. In each metric, the highest value is bolded. For DHMC, recall is represented from 30\% due to the limited number of test cases. A dagger $\dagger$ indicates that the calibration methods require extra inference steps for model calibration training.}
\label{tab:tab1}
\end{table*}

\section{Experiment}
\subsection{Experiment Settings}
\noindent\textbf{Model Calibration Methods}
We employ various methods to compare model calibration performance. Label smoothing~\cite{szegedy2016rethinking} regularizes the model by replacing hard labels with softened labels. Temperature scaling~\cite{guo2017calibration} recalibrates probabilities by dividing the logits by a learned scalar parameter before the softmax function. Monte Carlo (MC) dropout~\cite{gal2016dropout} estimates the uncertainty by multiple forward passes with different dropout masks at inference time. The estimated value of the MC dropout is obtained with 10 iterations with $p=0.5$. The deep ensemble~\cite{lakshminarayanan2017simple} combines predictions from multiple independently trained models to produce more reliable calibrated probabilities. We used 10 independent networks for the estimation of the ensemble. UDLS~\cite{park2025uncertainty} uses patch feature dropout to measure entropy, yielding sample-wise softened labels for further training.
\vspace{0.15cm}
\\
\noindent\textbf{Comparison Models for MRMIL}
We utilize state-of-the-art MRMIL architectures, each uniquely utilizing multi-resolution WSIs. DS-MIL~\cite{li2021dual}, representing an early transition from single to multi-resolution analysis, concatenates patches from an image pyramid with identical receptive fields. HAG-MIL~\cite{xiong2023diagnose} utilizes a progressive feature forwarding mechanism, comparing attention scores from coarse to fine resolutions, allowing a sequential exploration of structural information. Godson et al.~\cite{godson2023multi} construct a graph with various resolutions of patches, which is then compressed into a single representation using graph neural network aggregation and pooling.
\vspace{0.15cm}
\\
\noindent\textbf{Dataset}
We used three different public WSI datasets to examine the generalizability of UFC-MIL to diverse pathology types. The CAMELYON16~\cite{bejnordi2017diagnostic} dataset comprises 400 multi-resolution WSIs of hematoxylin and eosin (H\&E) stained lymph node sections. Department of Pathology and Laboratory Medicine at Dartmouth-Hitchcock Medical Center (DHMC)~\cite{wei2019pathologist} dataset comprises 143 H\&E-stained WSIs of lung adenocarcinoma. We preprocessed these data into binary classes of favorable (\textit{i}.\textit{e}., Lepidic, Acinar, Papillary) and poor (\textit{i}.\textit{e}., Micropapillary, Solid) prognosis cases for use. Since DHMC has no pre-defined splits, we divide it into 5-folds for our experiments. We further employed the Early Breast Cancer Core-Needle Biopsy WSI (BCNB) dataset~\cite{xu2021predicting}, comprising 1,058 cases, to classify Estrogen Receptor (ER) status, an important indicator for patient prognosis. For these datasets, we selected $\times256$ size patches from the 2, 1, and 0.5 Microns Per Pixel (MPP) of WSIs based on the Otsu algorithm~\cite{otsu1975threshold}. The patches were manually extracted by a pre-trained encoder~\cite{kang2023benchmarking}.
\vspace{0.15cm}
\\
\noindent\textbf{Implementation Details}
We set the hyperparameters $(\delta,\alpha)$ to $(0.49, 0.1)$, for which the sensitivity analysis is presented in the supplementary material. The model was optimized using Adam~\cite{kingma2014adam} with a learning rate of $1e-4$ and $\beta=(0.9,0.999)$, which is annealed to 0 over the total training epochs using a cosine scheduler~\cite{loshchilov2016sgdr}. All experiments were carried out on a single $\text{NVIDIA}^\circledR$ A6000. We performed all experiments multiple times with optimized hyperparameters for each method.
\vspace{0.15cm}
\\
\noindent\textbf{Evaluation Metrics}
We employ the expected calibration error (ECE) for calibration performance, which quantifies the difference between a model's predicted probabilities and its true accuracy across different confidence bins:
\begin{equation}
    ECE=\sum_{m=1}^M\frac{|B_m|}{N}{\mid}Acc(B_m)-Conf(B_m)\mid
\end{equation}
where $B_m$ is the set of samples in bin $m$ and $N$ is the total sample count. Accuracy $Acc(\cdot)$ and confidence $Conf(\cdot)$ are defined as the average values of the samples in that bin.
\begin{equation}
    Acc(B_m)=\frac{1}{|B_m|}\sum_{i\in{B_m}}\mathbb{1}(\hat{Y}_i=Y_i)\\
\end{equation}
\begin{equation}
    Conf(B_m)=\frac{1}{|B_m|}\sum_{i\in{B_m}}\hat{p}_i\left[\operatorname{argmax}_c\hat{p}_i[c]\right].
\end{equation}
We measure the recall of the top-$k$\% most confident predictions, which is denoted $\text{R@}k\text{\%}$, to gauge the reliability of the model, indicating the trustworthiness of its outputs.

\subsection{Model Calibration Results}
\subsubsection{Quantitative Performance}
We present the results of the quantitative evaluation for various calibration methods in Tab.~\ref{tab:tab1}.
In uncalibrated results, HAG-MIL and DS-MIL exhibit a high ECE, indicating that approaches focusing solely on classification accuracy are insufficient for enhancing model reliability.
MRMILs generally show calibration improvements with temperature scaling. Although some models experience a trade-off in ECE, they show better recall performance. Meanwhile, HAG-MIL exhibits a recall collapse in DHMC, revealing the vulnerability of models not optimized globally.
Label smoothing shows improved ECE and accuracy across all models and datasets, demonstrating the exceptional effectiveness of simply softening targets in calibrating model estimations.
M.C. dropout reduces the calibration error via repeated probabilistic inference. In particular, improvements in recall scores demonstrate its promise as a more reliable baseline within MRMIL. However, its iterative inference poses a challenge for practical use. Moreover, it requires a sacrifice in accuracy and recall in BCNB dataset, with the tendency to over-predict the positive class.
The deep ensemble method shows a conservative and stable performance improvement, which occurs across all metrics, but fail to introduce notable changes.
In the CAMELYON16 and DHMC datasets, UDLS exhibits performance degradation, increasing ECE, and lowering accuracy. In the BCNB data set, it improves the ECE but with a marginal or negative impact on other metrics, suggesting that the patch feature dropout employed to estimate the sample entropy is insufficient to build a high-quality target within a multiresolution context, a point not previously discussed for this approach.
The calibration that utilizes the multiple outputs of UFC-MIL shows improved ECE performance without requiring additional inference for UFC-MIL$^\bigstar$. In particular, it shows gains in both calibration and accuracy.

\begin{figure*}
    \centering 

    \begin{minipage}[c]{0.05\linewidth} 
        \textbf{(a)}
    \end{minipage}%
    \begin{minipage}[c]{0.94\linewidth} 
        \centering 
        \includegraphics[width=\linewidth]{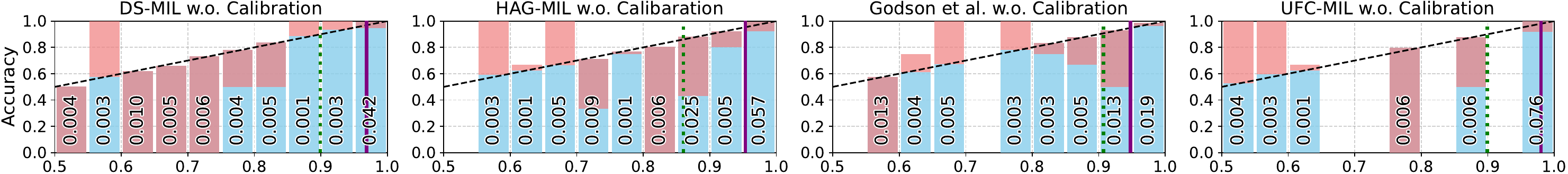} 
    \end{minipage}\par\vspace{1em} 

    \begin{minipage}[c]{0.05\linewidth}\label{fig:fig3-a}
        \textbf{(b)}
    \end{minipage}%
    \begin{minipage}[c]{0.94\linewidth}
        \centering
        \includegraphics[width=\linewidth]{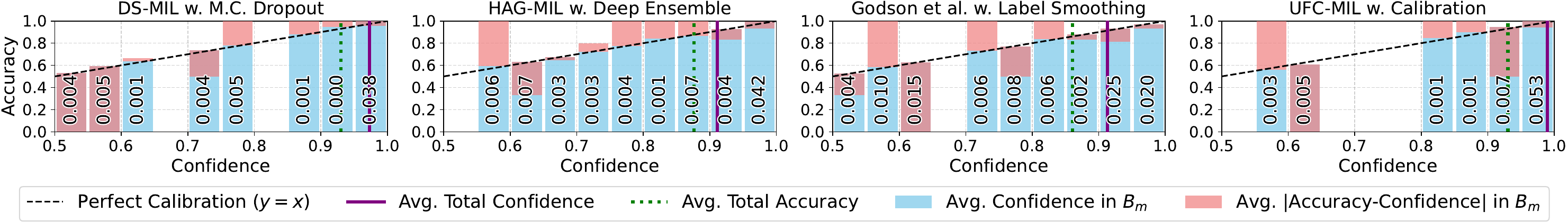}
    \end{minipage}



    \caption{Reliability diagrams on CAMELYON16. We plot histograms comparing uncalibrated models \textbf{(a)} with methods achieving the best ECE \textbf{(b)} for each. Analysis on DHMC and BCNB is presented in the supplementary material.}
    \label{fig:reliability}
\end{figure*}

\begin{table}[]
\centering
\resizebox{0.95\columnwidth}{!}{%
\begin{tabular}{l|cc|cc}
\hline
MIL                                  & \begin{tabular}[c]{@{}c@{}}Multi-\\ Resolution\end{tabular} & \begin{tabular}[c]{@{}c@{}}Feature\\ Extractor\\ Fine-Tuning\end{tabular} & AUC $\uparrow$  & Accuarcy $\uparrow$  \\ \hline
AB-MIL~\cite{ilse2018attention}      & \xmark                                                      & \xmark                                                                    & 0.865          & 0.845          \\
AB-MIL-MS~\cite{ilse2018attention}   & \cmark                                                      & \xmark                                                                    & 0.887          & 0.876          \\
DTFD-MIL(AFS)~\cite{zhang2022dtfd}   & \xmark                                                      & \xmark                                                                    & 0.946          & 0.908          \\
DS-MIL-Single~\cite{li2021dual}             & \xmark                                                      & \xmark                                                                    & 0.894          & 0.868          \\
DS-MIL~\cite{li2021dual}          & \cmark                                                      & \xmark                                                                    & 0.924          & 0.909          \\
TransMIL~\cite{shao2021transmil}     & \xmark                                                      & \xmark                                                                    & 0.942          & 0.883          \\
HIPT~\cite{chen2022scaling}          & \cmark                                                      & \cmark                                                                    & 0.951          & 0.890          \\
HAG-MIL~\cite{xiong2023diagnose}     & \cmark                                                      & \xmark                                                                    & 0.877          & 0.847          \\
Godson et al.~\cite{godson2023multi} & \cmark                                                      & \xmark                                                                    & 0.952          & 0.894          \\
DAS-MIL~\cite{bontempo2023mil}       & \cmark                                                      & \xmark                                                                    & 0.928          & 0.906          \\
DAS-MIL~\cite{bontempo2023mil}       & \cmark                                                      & \cmark                                                                    & \textbf{0.973} & 0.945          \\
Snuffy~\cite{jafarinia2024snuffy}    & \cmark                                                      & \cmark                                                                    & 0.970          & \textbf{0.952} \\ \hline
UFC-MIL                              & \cmark                                                      & \xmark                                                                    & 0.952          & 0.917          \\
UFC-MIL$^\bigstar$                   & \cmark                                                      & \xmark                                                                    & 0.964*          & 0.941*          \\ \hline
\end{tabular}%
}
\caption{Classification performance comparison on CAMELYON16 with the state-of-the-arts. *Among the models that do not require feature extractor fine-tuning, UFC-MIL$^\bigstar$ shows the highest performance.} 
\label{tab:tab-sota}
\end{table}

\subsubsection{Qualitave Analysis}
The reliability histogram in Fig.~\ref{fig:reliability} qualitatively illustrates the ECE for each model and the calibration method. As depicted in Fig.~\ref{fig:reliability}\textbf{(a)}, DS-MIL, HAG-MIL, and Godson et al. all produce prediction probabilities throughout the range. However, they do not align with their actual accuracy, indicating ill-calibrations. In contrast, UFC-MIL makes predictions by distinguishing between certain and uncertain cases.
For M.C. dropout applied to DS-MIL, predictions were binarized, but the accuracy for uncertain cases is low. The deep ensemble contributed to narrowing the gap between average confidence and accuracy, but binarized label smoothing predictions did not narrow the gap between confidence and accuracy.
Our proposed approach distinctly partitions prediction confidences. UFC-MIL with calibration closes the gap between confidence and accuracy, while retaining low confidence for difficult samples, ensuring that users can confidently decide whether to trust the model output.

\begin{figure}[t]
  \centering
   \includegraphics[width=0.9\linewidth]{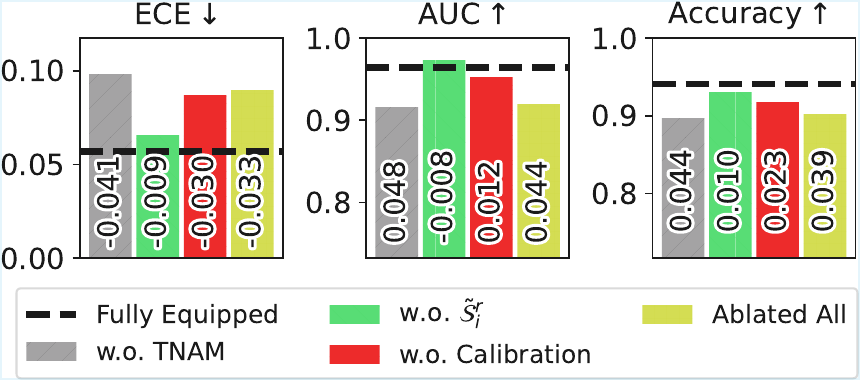}
   \caption{Performance with all proposed methods is shown by a dashed line (\ie, UFC-MIL$^\bigstar$), with the difference from each ablation indicated above the bars.}
   \label{fig:ablation}
\end{figure}

\subsection{Classification Performance}
Tab.~\ref{tab:tab-sota} provides a comparison of the performance of UFC-MIL with the milestone MILs on CAMELYON16. MRMIL generally exhibits improved performance compared to single-resolution MIL, as is clearly observable from the results on different resolution strategies within AB-MIL and DS-MIL. It should be noted that the state-of-the-art models~\cite{jafarinia2024snuffy,bontempo2023mil} achieve the best performance by jointly training their feature extractors on the target data. UFC-MIL models demonstrate superior performance among the models that do not require this fine-tuning. Furthermore, even without the advantage of feature extractor tuning, our calibrated UFC-MIL$^\bigstar$ achieves a classification performance that is comparable to that of current state-of-the-art models.


\subsection{Ablation Study}
We ablate TNAM and calibration training to verify the impact of each component on performance (Fig.~\ref{fig:ablation}) using CAMELYON16. Ablation of TNAM consistently produced substantial performance degradation, indicating that the absence of spatial information from TNAM results in a less accurate $\hat{\mathbf{p}}$, which consequently impacts calibration training. The proposed calibration component affected performance across all metrics. Not only did it improve ECE, which was the objective of training, but also improved classification performance. In particular, incorporating $\tilde{\mathcal{S}}$ into the smoothing factor $\epsilon$ positively impacted performance across multiple metrics, further indicating that the variance of entropy should also be considered in the sample uncertainty.

\subsection{Visualizing the Diagnostic Process}
\begin{figure}[t]
  \centering
   \includegraphics[width=1\columnwidth]{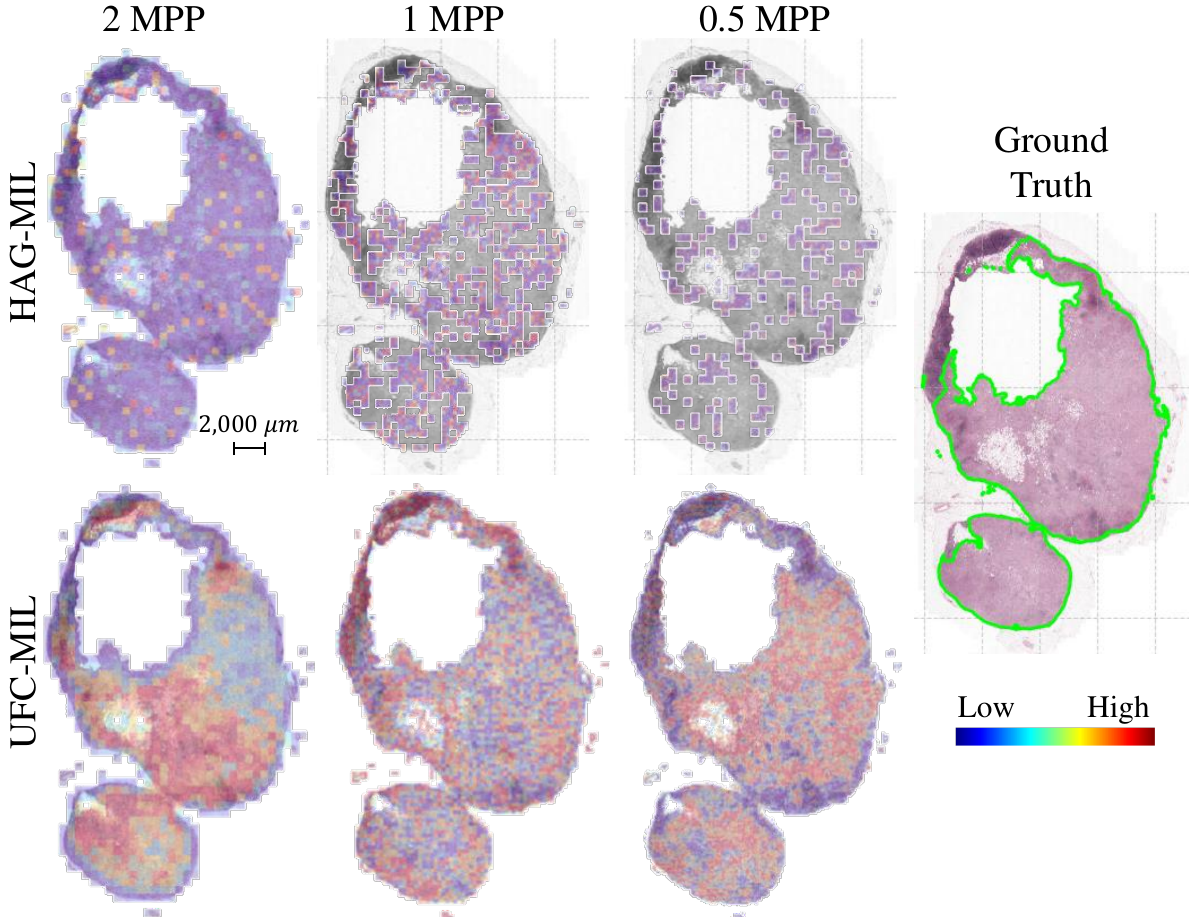}
   \caption{Illustration of attention map versus uncertainty map. In the attention map of HAG-MIL~\cite{xiong2023diagnose}, patches with low attention scores that were dropped during the zooming process are shown in grayscale, which the fine-grained model had no opportunity to observe. Additional cases are found in the supplementary material.}
   \label{fig:5}
\end{figure}
Fig.~\ref{fig:5} illustrates how UFC-MIL progressively resolves uncertainty through magnification, following the diagnostic behavior of pathologists. The figure includes HAG-MIL attention maps and UFC-MIL uncertainty maps from the coarsest (2 MPP) to the finest (0.5 MPP). At 2 MPP, HAG-MIL was motivated to attend to the lesions, but did not hit the area correctly. HAG-MIL's discrete and dropping inference strategy on low-attention patches led to an exponential decrease in the information at finer resolutions. In contrast, UFC-MIL emulated a human pathologist's workflow by continuously observing all patches from coarse to fine in an end-to-end manner. Interestingly, regions that UFC-MIL identified as uncertain at the coarse resolution level are often mitigated at the finer resolution level. The model then identified and focused on new uncertain areas that were not observable at the coarser resolution. This process reflects how an expert concentrates on uncertain regions, resolves uncertainty through magnification, and identifies new uncertainty areas at the finer level.

\subsection{Comparison with Various Position Strategies}
\begin{table}[t]
\resizebox{0.95\columnwidth}{!}{%
\begin{tabular}{c|c|cc|cc}
\hline
\multirow{2}{*}{Method}              & \multirow{2}{*}{Learnable} & \multicolumn{2}{c|}{CAMELYON16~\cite{bejnordi2017diagnostic}} & \multicolumn{2}{c}{DHMC~\cite{wei2019pathologist}} \\ \cline{3-6} 
                                     &                            & AUC $\uparrow$              & Accuracy $\uparrow$             & AUC $\uparrow$        & Accuracy $\uparrow$        \\ \hline
-                                    & -                          & 0.915                       & 0.896                           & 0.829                 & 0.762                      \\
Absolute~\cite{vaswani2017attention} & \xmark                     & 0.921                       & 0.894                           & 0.830                 & 0.769                      \\
Absolute~\cite{dosovitskiy2020image} & \cmark                     & 0.932                       & 0.896                           & 0.828                 & 0.773                      \\
PPEG~\cite{shao2021transmil}         & \cmark                     & 0.951                       & 0.902                           & 0.833                 & 0.782                      \\
Relative~\cite{su2024roformer}       & \cmark                     & 0.856                       & 0.868                           & 0.820                 & 0.759                      \\ \hline
TNAM                                 & \cmark                     & \textbf{0.952}              & \textbf{0.917}                  & \textbf{0.836}        & \textbf{0.793}             \\ \hline
\end{tabular}%
}
\caption{Various positional strategies and their results.}
\label{tab:position}
\end{table}

To examine the influence of positional information on performance, we integrate various strategies into UFC-MIL (Tab.~\ref{tab:position}) and compare them. Rule-based absolute strategy~\cite{vaswani2017attention} was not suitable for the MIL task. Its absolute position assumption did not align with the irregular shapes of pathological tissues and random instance bags. This tendency was similar to the learnable alternative~\cite{dosovitskiy2020image}. The PPEG module~\cite{shao2021transmil} shows improved performance, demonstrating that learnable convolutional operations at absolute positions can also contribute to MRMIL. The relative strategy~\cite{su2024roformer} could not adequately handle complex patch relationships in MRMIL, as its 2D rotative nature proved inadequate for a hierarchical multiresolution context. The superior performance of TNAM indicates that it effectively manages rotation-invariant information and aggregate it through learnable weights.
\section{Conclusion}

Inspired by the top-down zooming behaviors of pathologists dealing with multiple resolution images, we propose UFC-MIL, which emphasizes the importance of handling uncertain areas in a systematic way. Its structure reflects the spatially invariant characteristics of pathological images and allows uncertain patches to be passed to the finer resolution via differentiable operations. Moreover, we highlight the model calibration issue, a previously overlooked aspect in MRMIL. Along with the PW loss that allows patch-level predictions, we propose SRLS, an inference-free calibration training approach that uses multiple outputs. Comparisons with various calibration methods reveal that UFC-MIL$^\bigstar$, utilizing PW and SRLS, achieves superior calibration performance, significantly bringing MIL closer to practical trustworthiness for clinical users. Furthermore, UFC-MIL produces classification performance comparable to that of state-of-the-art MRMIL architectures. Further experiments offer deeper insights into the underlying mechanisms of the proposed model. Our work broadens the scope of MRMIL by shifting its focus to the issue of uncertainty to deliver a well-calibrated model, which is a critical attribute for clinical applications.

\section*{Acknowledgment}
This work was supported by the National Research Foundation of Korea (NRF) grant funded by the Korea government (MSIT) (No.RS-2022-NR068758)
{\small
\bibliographystyle{ieee_fullname}
\bibliography{0.Main.bib}

@article{article,
author = {Bychkov, Andrey and Schubert, Michael},
year = {2023},
month = {02},
pages = {18-27},
title = {Constant Demand, Patchy Supply},
volume = {88}
}

@article{bray2021ever,
  title={The ever-increasing importance of cancer as a leading cause of premature death worldwide},
  author={Bray, Freddie and Laversanne, Mathieu and Weiderpass, Elisabete and Soerjomataram, Isabelle},
  journal={Cancer},
  volume={127},
  number={16},
  pages={3029--3030},
  year={2021},
  publisher={Wiley Online Library}
}

@article{zhang2025patches,
  title={From patches to WSIs: A systematic review of deep Multiple Instance Learning in computational pathology},
  author={Zhang, Yuchen and Gao, Zeyu and He, Kai and Li, Chen and Mao, Rui},
  journal={Information Fusion},
  pages={103027},
  year={2025},
  publisher={Elsevier}
}

@article{gadermayr2024multiple,
  title={Multiple instance learning for digital pathology: A review of the state-of-the-art, limitations \& future potential},
  author={Gadermayr, Michael and Tschuchnig, Maximilian},
  journal={Computerized Medical Imaging and Graphics},
  volume={112},
  pages={102337},
  year={2024},
  publisher={Elsevier}
}

@article{shao2021transmil,
  title={Transmil: Transformer based correlated multiple instance learning for whole slide image classification},
  author={Shao, Zhuchen and Bian, Hao and Chen, Yang and Wang, Yifeng and Zhang, Jian and Ji, Xiangyang and others},
  journal={Advances in neural information processing systems},
  volume={34},
  pages={2136--2147},
  year={2021}
}

@inproceedings{zhang2022dtfd,
  title={Dtfd-mil: Double-tier feature distillation multiple instance learning for histopathology whole slide image classification},
  author={Zhang, Hongrun and Meng, Yanda and Zhao, Yitian and Qiao, Yihong and Yang, Xiaoyun and Coupland, Sarah E and Zheng, Yalin},
  booktitle={Proceedings of the IEEE/CVF conference on computer vision and pattern recognition},
  pages={18802--18812},
  year={2022}
}

@inproceedings{ilse2018attention,
  title={Attention-based deep multiple instance learning},
  author={Ilse, Maximilian and Tomczak, Jakub and Welling, Max},
  booktitle={International conference on machine learning},
  pages={2127--2136},
  year={2018},
  organization={PMLR}
}

@article{lu2021data,
  title={Data-efficient and weakly supervised computational pathology on whole-slide images},
  author={Lu, Ming Y and Williamson, Drew FK and Chen, Tiffany Y and Chen, Richard J and Barbieri, Matteo and Mahmood, Faisal},
  journal={Nature biomedical engineering},
  volume={5},
  number={6},
  pages={555--570},
  year={2021},
  publisher={Nature Publishing Group UK London}
}

@inproceedings{hou2022h,
  title={H\^{} 2-MIL: exploring hierarchical representation with heterogeneous multiple instance learning for whole slide image analysis},
  author={Hou, Wentai and Yu, Lequan and Lin, Chengxuan and Huang, Helong and Yu, Rongshan and Qin, Jing and Wang, Liansheng},
  booktitle={Proceedings of the AAAI conference on artificial intelligence},
  volume={36},
  number={1},
  pages={933--941},
  year={2022}
}

@inproceedings{li2021dual,
  title={Dual-stream multiple instance learning network for whole slide image classification with self-supervised contrastive learning},
  author={Li, Bin and Li, Yin and Eliceiri, Kevin W},
  booktitle={Proceedings of the IEEE/CVF conference on computer vision and pattern recognition},
  pages={14318--14328},
  year={2021}
}

@article{clevert2015fast,
  title={Fast and accurate deep network learning by exponential linear units (elus)},
  author={Clevert, Djork-Arn{\'e} and Unterthiner, Thomas and Hochreiter, Sepp},
  journal={arXiv preprint arXiv:1511.07289},
  year={2015}
}

@inproceedings{jafarinia2024snuffy,
  title={Snuffy: Efficient Whole Slide Image Classifier},
  author={Jafarinia, Hossein and Alipanah, Alireza and Razavi, Saeed and Mirzaie, Nahal and Rohban, Mohammad Hossein},
  booktitle={European Conference on Computer Vision},
  pages={243--260},
  year={2024},
  organization={Springer}
}

@inproceedings{chen2022scaling,
  title={Scaling vision transformers to gigapixel images via hierarchical self-supervised learning},
  author={Chen, Richard J and Chen, Chengkuan and Li, Yicong and Chen, Tiffany Y and Trister, Andrew D and Krishnan, Rahul G and Mahmood, Faisal},
  booktitle={Proceedings of the IEEE/CVF conference on computer vision and pattern recognition},
  pages={16144--16155},
  year={2022}
}

@article{huang2023cross,
  title={Cross-scale fusion transformer for histopathological image classification},
  author={Huang, Sheng-Kai and Yu, Yu-Ting and Huang, Chun-Rong and Cheng, Hsiu-Chi},
  journal={IEEE Journal of Biomedical and Health Informatics},
  volume={28},
  number={1},
  pages={297--308},
  year={2023},
  publisher={IEEE}
}

@article{xiong2023diagnose,
  title={Diagnose like a pathologist: Transformer-enabled hierarchical attention-guided multiple instance learning for whole slide image classification},
  author={Xiong, Conghao and Chen, Hao and Sung, Joseph JY and King, Irwin},
  journal={arXiv preprint arXiv:2301.08125},
  year={2023}
}

@inproceedings{bontempo2023mil,
  title={DAS-MIL: distilling across scales for MIL classification of histological WSIs},
  author={Bontempo, Gianpaolo and Porrello, Angelo and Bolelli, Federico and Calderara, Simone and Ficarra, Elisa},
  booktitle={International conference on medical image computing and computer-assisted intervention},
  pages={248--258},
  year={2023},
  organization={Springer}
}

@inproceedings{guo2017calibration,
  title={On calibration of modern neural networks},
  author={Guo, Chuan and Pleiss, Geoff and Sun, Yu and Weinberger, Kilian Q},
  booktitle={International conference on machine learning},
  pages={1321--1330},
  year={2017},
  organization={PMLR}
}

@inproceedings{xiong2021nystromformer,
  title={Nystr{\"o}mformer: A nystr{\"o}m-based algorithm for approximating self-attention},
  author={Xiong, Yunyang and Zeng, Zhanpeng and Chakraborty, Rudrasis and Tan, Mingxing and Fung, Glenn and Li, Yin and Singh, Vikas},
  booktitle={Proceedings of the AAAI conference on artificial intelligence},
  volume={35},
  number={16},
  pages={14138--14148},
  year={2021}
}

@article{quellec2012multiple,
  title={A multiple-instance learning framework for diabetic retinopathy screening},
  author={Quellec, Gw{\'e}nol{\'e} and Lamard, Mathieu and Abr{\`a}moff, Michael D and Decenci{\`e}re, Etienne and Lay, Bruno and Erginay, Ali and Cochener, B{\'e}atrice and Cazuguel, Guy},
  journal={Medical image analysis},
  volume={16},
  number={6},
  pages={1228--1240},
  year={2012},
  publisher={Elsevier}
}

@inproceedings{park2025uncertainty,
  title={Uncertainty-based Data-wise Label Smoothing for Calibrating Multiple Instance Learning in Histopathology Image Classification},
  author={Park, Hyeongmin and Hong, Sungrae and Song, Chanjae and Kim, Jongwoo and Yi, Mun Yong},
  booktitle={2025 IEEE/CVF Winter Conference on Applications of Computer Vision (WACV)},
  pages={599--608},
  year={2025},
  organization={IEEE}
}

@article{eloy2023artificial,
  title={Artificial intelligence--assisted cancer diagnosis improves the efficiency of pathologists in prostatic biopsies},
  author={Eloy, Catarina and Marques, Ana and Pinto, Jo{\~a}o and Pinheiro, Jorge and Campelos, Sofia and Curado, M{\'o}nica and Vale, Jo{\~a}o and Pol{\'o}nia, Ant{\'o}nio},
  journal={Virchows Archiv},
  volume={482},
  number={3},
  pages={595--604},
  year={2023},
  publisher={Springer}
}

@article{dolezal2022uncertainty,
  title={Uncertainty-informed deep learning models enable high-confidence predictions for digital histopathology},
  author={Dolezal, James M and Srisuwananukorn, Andrew and Karpeyev, Dmitry and Ramesh, Siddhi and Kochanny, Sara and Cody, Brittany and Mansfield, Aaron S and Rakshit, Sagar and Bansal, Radhika and Bois, Melanie C and others},
  journal={Nature communications},
  volume={13},
  number={1},
  pages={6572},
  year={2022},
  publisher={Nature Publishing Group UK London}
}

@article{raab2005clinical,
  title={Clinical impact and frequency of anatomic pathology errors in cancer diagnoses},
  author={Raab, Stephen S and Grzybicki, Dana Marie and Janosky, Janine E and Zarbo, Richard J and Meier, Frederick A and Jensen, Chris and Geyer, Stanley J},
  journal={Cancer: Interdisciplinary International Journal of the American Cancer Society},
  volume={104},
  number={10},
  pages={2205--2213},
  year={2005},
  publisher={Wiley Online Library}
}

@article{ghezloo2022analysis,
  title={An analysis of pathologists’ viewing processes as they diagnose whole slide digital images},
  author={Ghezloo, Fatemeh and Wang, Pin-Chieh and Kerr, Kathleen F and Bruny{\'e}, Tad T and Drew, Trafton and Chang, Oliver H and Reisch, Lisa M and Shapiro, Linda G and Elmore, Joann G},
  journal={Journal of Pathology Informatics},
  volume={13},
  pages={100104},
  year={2022},
  publisher={Elsevier}
}

@article{brunye2017accuracy,
  title={Accuracy is in the eyes of the pathologist: the visual interpretive process and diagnostic accuracy with digital whole slide images},
  author={Bruny{\'e}, Tad T and Mercan, Ezgi and Weaver, Donald L and Elmore, Joann G},
  journal={Journal of biomedical informatics},
  volume={66},
  pages={171--179},
  year={2017},
  publisher={Elsevier}
}

@inproceedings{ramon2000multi,
  title={Multi instance neural networks},
  author={Ramon, Jan and De Raedt, Luc},
  booktitle={Proceedings of the ICML-2000 workshop on attribute-value and relational learning},
  pages={53--60},
  year={2000}
}

@article{dosovitskiy2020image,
  title={An image is worth 16x16 words: Transformers for image recognition at scale},
  author={Dosovitskiy, Alexey and Beyer, Lucas and Kolesnikov, Alexander and Weissenborn, Dirk and Zhai, Xiaohua and Unterthiner, Thomas and Dehghani, Mostafa and Minderer, Matthias and Heigold, Georg and Gelly, Sylvain and others},
  journal={arXiv preprint arXiv:2010.11929},
  year={2020}
}

@article{bejnordi2017diagnostic,
  title={Diagnostic assessment of deep learning algorithms for detection of lymph node metastases in women with breast cancer},
  author={Bejnordi, Babak Ehteshami and Veta, Mitko and Van Diest, Paul Johannes and Van Ginneken, Bram and Karssemeijer, Nico and Litjens, Geert and Van Der Laak, Jeroen AWM and Hermsen, Meyke and Manson, Quirine F and Balkenhol, Maschenka and others},
  journal={Jama},
  volume={318},
  number={22},
  pages={2199--2210},
  year={2017},
  publisher={American Medical Association}
}

@inproceedings{blundell2015weight,
  title={Weight uncertainty in neural network},
  author={Blundell, Charles and Cornebise, Julien and Kavukcuoglu, Koray and Wierstra, Daan},
  booktitle={International conference on machine learning},
  pages={1613--1622},
  year={2015},
  organization={PMLR}
}

@inproceedings{gal2016dropout,
  title={Dropout as a bayesian approximation: Representing model uncertainty in deep learning},
  author={Gal, Yarin and Ghahramani, Zoubin},
  booktitle={international conference on machine learning},
  pages={1050--1059},
  year={2016},
  organization={PMLR}
}

@article{lakshminarayanan2017simple,
  title={Simple and scalable predictive uncertainty estimation using deep ensembles},
  author={Lakshminarayanan, Balaji and Pritzel, Alexander and Blundell, Charles},
  journal={Advances in neural information processing systems},
  volume={30},
  year={2017}
}

@inproceedings{zadrozny2001obtaining,
  title={Obtaining calibrated probability estimates from decision trees and naive bayesian classifiers},
  author={Zadrozny, Bianca and Elkan, Charles},
  booktitle={Icml},
  volume={1},
  number={05},
  year={2001}
}

@article{pereyra2017regularizing,
  title={Regularizing neural networks by penalizing confident output distributions},
  author={Pereyra, Gabriel and Tucker, George and Chorowski, Jan and Kaiser, {\L}ukasz and Hinton, Geoffrey},
  journal={arXiv preprint arXiv:1701.06548},
  year={2017}
}

@article{muller2019does,
  title={When does label smoothing help?},
  author={M{\"u}ller, Rafael and Kornblith, Simon and Hinton, Geoffrey E},
  journal={Advances in neural information processing systems},
  volume={32},
  year={2019}
}

@inproceedings{rymarczyk2022protomil,
  title={Protomil: Multiple instance learning with prototypical parts for whole-slide image classification},
  author={Rymarczyk, Dawid and Pardyl, Adam and Kraus, Jaros{\l}aw and Kaczy{\'n}ska, Aneta and Skomorowski, Marek and Zieli{\'n}ski, Bartosz},
  booktitle={Joint European conference on machine learning and knowledge discovery in databases},
  pages={421--436},
  year={2022},
  organization={Springer}
}

@article{hense2024xmil,
  title={XMIL: Insightful explanations for multiple instance learning in histopathology},
  author={Hense, Julius and Jamshidi Idaji, Mina and Eberle, Oliver and Schnake, Thomas and Dippel, Jonas and Ciernik, Laure and Buchstab, Oliver and Mock, Andreas and Klauschen, Frederick and M{\"u}ller, Klaus-Robert},
  journal={Advances in Neural Information Processing Systems},
  volume={37},
  pages={8300--8328},
  year={2024}
}

@article{wang2023calibration,
  title={Calibration in deep learning: A survey of the state-of-the-art},
  author={Wang, Cheng},
  journal={arXiv preprint arXiv:2308.01222},
  year={2023}
}

@article{li2021localization,
  title={Localization with sampling-argmax},
  author={Li, Jiefeng and Chen, Tong and Shi, Ruiqi and Lou, Yujing and Li, Yong-Lu and Lu, Cewu},
  journal={Advances in Neural Information Processing Systems},
  volume={34},
  pages={27236--27248},
  year={2021}
}

@article{otsu1975threshold,
  title={A threshold selection method from gray-level histograms},
  author={Otsu, Nobuyuki and others},
  journal={Automatica},
  volume={11},
  number={285-296},
  pages={23--27},
  year={1975}
}

@article{loshchilov2016sgdr,
  title={Sgdr: Stochastic gradient descent with warm restarts},
  author={Loshchilov, Ilya and Hutter, Frank},
  journal={arXiv preprint arXiv:1608.03983},
  year={2016}
}

@inproceedings{godson2023multi,
  title={Multi-level Graph Representations of Melanoma Whole Slide Images for Identifying Immune Subgroups},
  author={Godson, Lucy and Alemi, Navid and Nsengimana, J{\'e}r{\'e}mie and Cook, Graham P and Clarke, Emily L and Treanor, Darren and Bishop, D Timothy and Newton-Bishop, Julia and Magee, Derek},
  booktitle={International Conference on Medical Image Computing and Computer-Assisted Intervention},
  pages={85--96},
  year={2023},
  organization={Springer}
}

@article{wei2019pathologist,
  title={Pathologist-level classification of histologic patterns on resected lung adenocarcinoma slides with deep neural networks},
  author={Wei, Jason W and Tafe, Laura J and Linnik, Yevgeniy A and Vaickus, Louis J and Tomita, Naofumi and Hassanpour, Saeed},
  journal={Scientific reports},
  volume={9},
  number={1},
  pages={3358},
  year={2019},
  publisher={Nature Publishing Group UK London}
}

@inproceedings{kang2023benchmarking,
  title={Benchmarking self-supervised learning on diverse pathology datasets},
  author={Kang, Mingu and Song, Heon and Park, Seonwook and Yoo, Donggeun and Pereira, S{\'e}rgio},
  booktitle={Proceedings of the IEEE/CVF Conference on Computer Vision and Pattern Recognition},
  pages={3344--3354},
  year={2023}
}

@article{xu2021predicting,
title={Predicting Axillary Lymph Node Metastasis in Early Breast Cancer Using Deep Learning on Primary Tumor Biopsy Slides},
author={Xu, Feng and Zhu, Chuang and Tang, Wenqi and Wang, Ying and Zhang, Yu and Li, Jie and Jiang, Hongchuan and Shi, Zhongyue and Liu, Jun and Jin, Mulan},
journal={Frontiers in Oncology},
pages={4133},
year={2021},
publisher={Frontiers}
}

@inproceedings{szegedy2016rethinking,
  title={Rethinking the inception architecture for computer vision},
  author={Szegedy, Christian and Vanhoucke, Vincent and Ioffe, Sergey and Shlens, Jon and Wojna, Zbigniew},
  booktitle={Proceedings of the IEEE conference on computer vision and pattern recognition},
  pages={2818--2826},
  year={2016}
}

@article{kingma2014adam,
  title={Adam: A method for stochastic optimization},
  author={Kingma, Diederik P},
  journal={arXiv preprint arXiv:1412.6980},
  year={2014}
}

@article{vaswani2017attention,
  title={Attention is all you need},
  author={Vaswani, Ashish and Shazeer, Noam and Parmar, Niki and Uszkoreit, Jakob and Jones, Llion and Gomez, Aidan N and Kaiser, {\L}ukasz and Polosukhin, Illia},
  journal={Advances in neural information processing systems},
  volume={30},
  year={2017}
}

@article{su2024roformer,
  title={Roformer: Enhanced transformer with rotary position embedding},
  author={Su, Jianlin and Ahmed, Murtadha and Lu, Yu and Pan, Shengfeng and Bo, Wen and Liu, Yunfeng},
  journal={Neurocomputing},
  volume={568},
  pages={127063},
  year={2024},
  publisher={Elsevier}
}

@article{jang2016categorical,
  title={Categorical reparameterization with gumbel-softmax},
  author={Jang, Eric and Gu, Shixiang and Poole, Ben},
  journal={arXiv preprint arXiv:1611.01144},
  year={2016}
}

@String(AAAI = {AAAI})
}

\end{document}